\documentclass[lettersize,journal]{IEEEtran}
\usepackage{booktabs}
\usepackage{tabularx}
\usepackage{siunitx}

\usepackage{graphicx}
\usepackage{cite}
\usepackage{makecell}  
\usepackage{array}     
\usepackage{ragged2e}  
\usepackage{url}
\usepackage{colortbl}
\usepackage{multirow}
\usepackage{amssymb}
\usepackage{mathtools}
\usepackage{arydshln}
\usepackage{bbm}
\usepackage{xcolor}
\usepackage{booktabs}

\usepackage[linesnumbered,ruled,vlined]{algorithm2e}

\usepackage{hyperref}
\hypersetup{
    colorlinks=true,
    linkcolor=red,
    filecolor=magenta,      
    urlcolor=blue,
}

\definecolor{limegreen}{rgb}{0.2, 0.8, 0.2}
\definecolor{forestgreen}{rgb}{0.13, 0.55, 0.13}
\definecolor{greenhtml}{rgb}{0.0, 0.5, 0.0}
\definecolor{black}{rgb}{0.0, 0.0, 0.0}

\usepackage[font=small,labelfont=bf,tableposition=top]{caption}

\usepackage{blindtext}
\title{here title}

\let\oldtwocolumn\twocolumn
\renewcommand\twocolumn[1][]{%
    \oldtwocolumn[{#1}{
    \begin{center}
           \includegraphics[width=18cm]{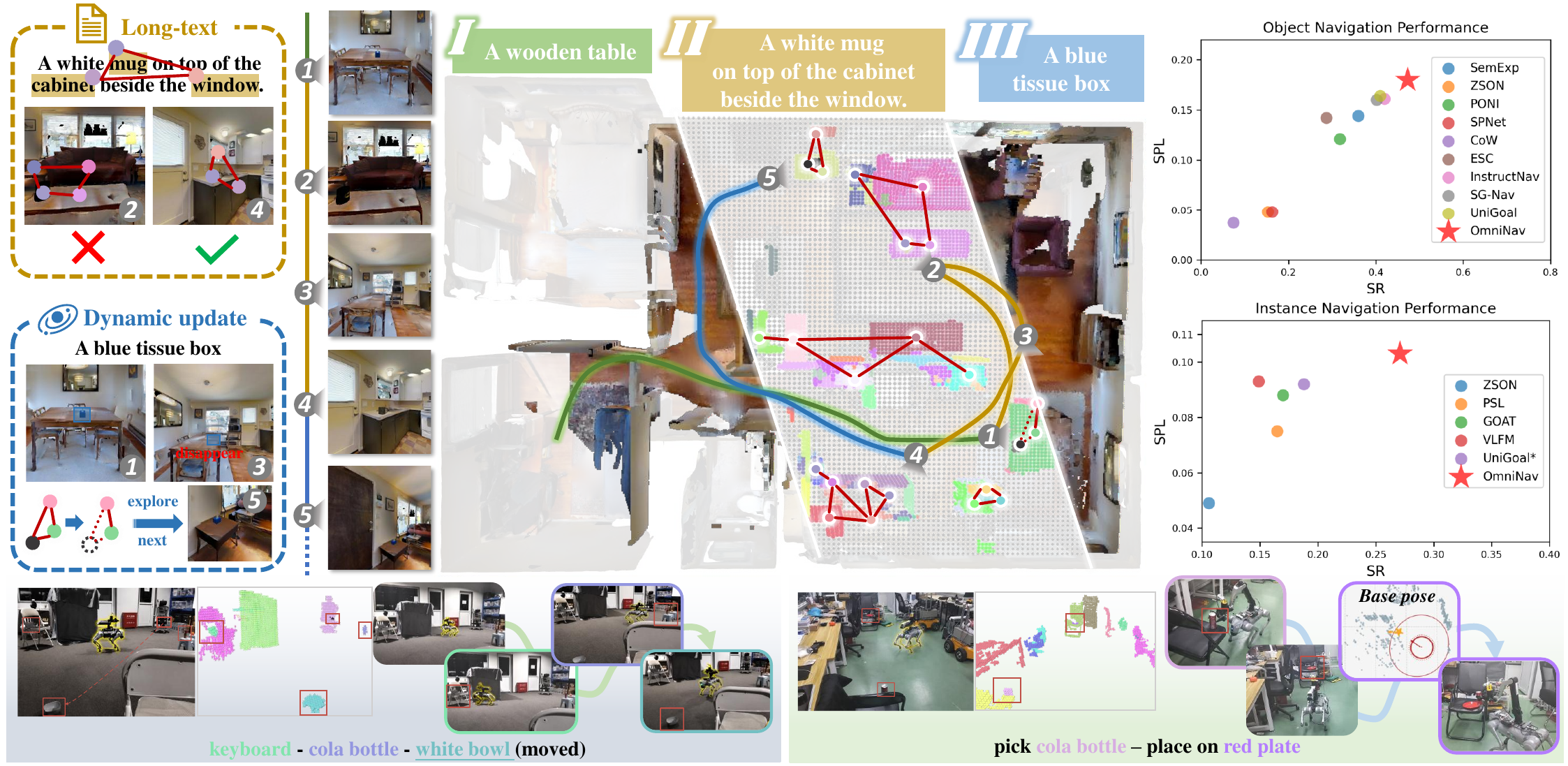}
           \captionof{figure}{Overview of \textbf{OmniNav}.
           OmniNav maintains a persistent object-centric scene memory, disambiguates fine-grained targets from complex language instructions, adapts its plan to dynamic scene changes, and performs reachability-aware closed-loop mobile manipulation. Together, these capabilities enable robust long-horizon target navigation and interaction in complex environments.
}
           \label{first}
        \end{center}
    }]
}

\begin{document}


\title
{	
\textbf{OmniNav}: Robust Long-Horizon Target Navigation in Dynamic Environments

\thanks{$^1$ Yujie Tang, Meiling Wang, Jinhao Jiang, Sibo Zuo, Yinan Deng, Xinyu Zhang, Yufeng Yue are with School of Automation, Beijing Institute of Technology, Beijing, China. (Corresponding Author: Yufeng Yue, yueyufeng@bit.edu.cn)}
}

\author{
Anonymous TRO submission
}

\author{Yujie Tang$^1$, Meiling Wang$^1$, Jinhao Jiang$^1$, Sibo Zuo$^1$, Yinan Deng$^1$, Xinyu Zhang$^1$, Yufeng Yue$^{*}$}

\maketitle

\begin{abstract}
Long-horizon target navigation requires a robot to sustain task execution
across evolving observations, decisions, and physical interactions.
This requires three coupled capabilities: 
maintaining valid scene memory, revising target beliefs under partial observability, and selecting interaction-feasible navigation endpoints. 
However, the state underlying each capability is only
conditionally valid: scene representations become stale when objects move
or disappear, unsuccessful searches alter beliefs over target locations,
and geometrically convenient endpoints may still be infeasible for
manipulation. To address these challenges, we present OmniNav, which formulates long-horizon navigation as continual inference over a factorized task state posterior coupling scene validity, target belief, and
interaction feasibility. For
representation, OmniNav incrementally constructs an updatable 3D object
scene memory, preventing stale scene evidence from propagating to
subsequent decisions. For exploration, it introduces an evidence-aware
Bayesian belief-revision mechanism that derives dependency-aware region
priors from semantic context, incorporates unsuccessful searches as
negative evidence, and updates them for posterior-guided frontier
selection. For interaction, OmniNav incorporates manipulation reachability
and collision constraints into navigation-endpoint selection and propagates
execution feedback through hierarchical closed-loop recovery. Extensive experiments demonstrate that OmniNav achieves the highest
success rates among the compared methods on semantic ObjectNav and
fine-grained instance navigation benchmarks,
remains robust to target relocation, and improves real-world pick-and-place
success from 53.3\% to 71.7\% over an adapted open-loop baseline.
The project page of OmniNav is available at \url{https://omni-nav.github.io/}.

\end{abstract}


\section{Introduction}
\label{sec:introduction}

Recent advances in embodied AI and mobile robotics have enabled robots to perform increasingly diverse navigation tasks in indoor environments, ranging from semantic object-goal navigation~\cite{chaplot2020object,ramakrishnan2022poni,gadre2023cows} and fine-grained language-guided target search~\cite{yin2025unigoal} to mobile manipulation~\cite{liu2024okrobot,liu2025dynamem}. Realistic robot deployment rarely consists of a short, isolated episode that terminates once a single goal is reached. In long-horizon tasks, a robot may need to interpret a natural-language instruction, locate candidate targets, distinguish the intended target from visually similar instances, approach it from an interaction-feasible base pose, execute a physical operation, and continue with subsequent tasks as the scene evolves. 
We therefore view real-world target navigation as a long-horizon closed-loop process,
where navigation not only supports target search but must also produce interaction-feasible outcomes and adapt to subsequent execution feedback.

\begin{table*}[htbp]
\centering
\scriptsize
\caption{Capability Comparison of Existing Embodied Navigation Frameworks.}
\label{tab_capacity_compare}

\renewcommand\arraystretch{1.25}
\setlength{\extrarowheight}{1.1pt}

\resizebox{\linewidth}{!}{
\setlength{\tabcolsep}{2mm}{
\begin{tabular}{c|c|c|c|c|c}

\Xhline{1.1pt}
\begin{tabular}[c]{@{}c@{}}Representative Works \end{tabular} 
& \begin{tabular}[c]{@{}c@{}}\textbf{Persistent} \\\textbf{Scene Memory}\\  \end{tabular} 
& \begin{tabular}[c]{@{}c@{}}\textbf{Online}\\ \textbf{Exploration}\\\end{tabular} 
& \begin{tabular}[c]{@{}c@{}}\textbf{Dynamic}\\ \textbf{Scene Adaptation}\\\end{tabular} 
& \begin{tabular}[c]{@{}c@{}}\textbf{Fine-Grained}\\ \textbf{Disambiguation}\\\end{tabular} 
& \begin{tabular}[c]{@{}c@{}}\textbf{Closed-Loop}\\ \textbf{Mobile Manipulation}\\\end{tabular}    
\\ 
\hline

Uni-NaVid \cite{zhang2024uni} (End-to-end Implicit Memory-based) 
& $\triangle$ & $\checkmark$  & $\times$ & $\triangle$ & $\times$   \\

VLFM \cite{yokoyama2024vlfm} (Feature Similarity Map-based) 
& $\triangle$ & $\checkmark$  & $\times$ & $\times$ & $\times$  \\

ApexNav \cite{zhang2025apexnav} (Target-Centric Semantic Map-based) 
& $\triangle$ & $\checkmark$  & $\times$ & $\times$ & $\times$  \\

RoboHop \cite{garg2024robohop} (Topological Graph-based) 
& $\triangle$ & $\checkmark$  & $\times$ & $\triangle$ & $\times$  \\

DualMap \cite{jiang2025dualmap} (Dynamic Object-Centric Map-based) 
& $\checkmark$ & $\checkmark$  & $\checkmark$ & $\times$ & $\times$  \\

DovSG \cite{yan2025dovsg} (Dynamic Object-Centric Map-based) 
& $\checkmark$ & $\times$  & $\checkmark$ & $\triangle$ & $\triangle$  \\

\hline
\textbf{OmniNav (Ours)} 
& $\checkmark$ & $\checkmark$ & $\checkmark$ & $\checkmark$ & $\checkmark$  
\\ 
\Xhline{1.1pt}

\end{tabular}
}
}

\vspace{0.5em}

\noindent
\begin{minipage}{\linewidth}
\footnotesize
\justifying
\sloppy
  \textbf{Note.} The table summarizes a representative selection of embodied
  navigation frameworks. The symbols $\checkmark$,
  $\triangle$, and $\times$ denote full, partial, and no explicit support,
  respectively. For \textbf{Persistent Scene Memory}, $\triangle$ indicates
  that historical scene information is retained through implicit or
  non-object-centric representations rather than an explicit persistent
  object-level memory. For \textbf{Dynamic Scene Adaptation}, $\checkmark$ requires explicit revision of stored scene representations under object or scene changes. For \textbf{Fine-Grained Disambiguation}, $\triangle$ denotes methods that partially support attribute-, relation-, or instruction-conditioned target reasoning but lack explicit candidate-level confirmation among visually similar instances. For \textbf{Closed-Loop Mobile
  Manipulation}, $\triangle$ indicates support for mobile manipulation with primarily
  open-loop execution or limited feedback-driven recovery.
\end{minipage}
\end{table*}

The central difficulty is that the task state supporting these stages remains
only conditionally valid: accumulated scene evidence can become stale, beliefs about where the target may be should change as unsuccessful searches accumulate, and a reachable navigation goal need not be feasible for physical interaction. 
Because these states are coupled, an outdated assumption at one stage can propagate to subsequent search, navigation, or interaction decisions.
Long-horizon navigation therefore requires continual revision of scene validity, target belief, and interaction feasibility.

Existing embodied frameworks cover complementary subsets of the capabilities required for long-horizon target navigation, as summarized in Table~\ref{tab_capacity_compare}. Navigation-oriented methods, including Uni-NaVid~\cite{zhang2024uni}, VLFM~\cite{yokoyama2024vlfm}, ApexNav~\cite{zhang2025apexnav}, and RoboHop~\cite{garg2024robohop}, primarily focus on online target-directed exploration using implicit memory,
feature-similarity maps, target-centric semantic maps, or topological representations. DualMap~\cite{jiang2025dualmap} further combines online exploration with persistent object-centric memory and dynamic scene adaptation. For physical interaction, DovSG~\cite{yan2025dovsg}
incorporates dynamic object-centric scene representations into mobile manipulation and provides partial support for fine-grained target reasoning. Together, these methods establish components for navigation,
scene representation, and physical interaction. 
However, these capabilities are optimized in isolation, providing limited mechanisms for evidence acquired at one stage to revise the state of another. 
Their joint treatment with cross-stage state revision in dynamic long-horizon execution remains insufficiently explored.

To address these limitations, we present \textbf{OmniNav}, a framework for
  robust long-horizon target navigation and interaction in dynamic indoor
  environments, as illustrated in Fig.~\ref{first}. Rather than treating scene
  mapping, target search, and physical interaction as independent stages,
  OmniNav formulates long-horizon execution as continual inference over a factorized task state posterior.
  From posed RGB-D observations, it constructs an open-vocabulary,
  object-centric 3D scene memory online and propagates new observations, search
  outcomes, and execution feedback to subsequent decisions, requiring neither a
  pre-built map nor task-specific training. Its design is organized around
  three coupled dimensions: maintaining scene validity, revising target belief,
  and ensuring interaction feasibility.

 \textbf{1) Scene validity: How can accumulated object memory remain
  trustworthy as the scene evolves?}
  Existing dynamic scene memories often revise movable objects through
carrier-, anchor-, or reassociation-based mechanisms~\cite{tang2025openin,tang2025openobject,jiang2025dualmap}.
Such mechanisms typically rely on stable reference structure or confident
object reassociation, which may become unreliable as the environment is
incrementally explored and modified. More fundamentally, when a previously
mapped object is not observed at its expected location, it may have been
relocated or removed, but it may also be occluded or temporarily missed.
This ambiguity makes both unconditional retention and immediate deletion
unreliable. OmniNav instead maintains an independent Bayesian persistence
state for every mapped instance, recursively updating it from supporting,
empty-space, and occlusion evidence and removing stale instances only when
their posterior validity falls below a threshold.

\textbf{2) Target belief: How should exploration revise where and
  what the target is?}
  Semantic exploration methods use visual--language similarity, object
  co-occurrence, or scene context to prioritize regions likely to contain the
  target~\cite{ramakrishnan2022poni,yokoyama2024vlfm,zhang2025apexnav,yin2025unigoal}.
  However, correlated context objects can produce overconfident semantic
  priors, and a region may remain highly ranked even after it has been
  extensively searched without revealing the target. Existing exploration
  coverage records where the robot has traveled but does not necessarily
  convert target-absent observations into target-dependent negative evidence.
  Moreover, region-level semantic relevance alone cannot distinguish the
  intended instance from visually similar candidates. OmniNav groups dependent
  context cues before probabilistic prior aggregation and then applies Bayesian
  updating to convert unsuccessful exploration into region-wise posterior
  target probabilities. Once a candidate is observed, it jointly verifies the
  candidate against accumulated scene context and the current egocentric VLM
  observation.

  \textbf{3) Interaction feasibility: How can a navigation outcome be
  converted into an executable physical operation?}
  Existing mobile manipulation systems often connect navigation and
physical execution sequentially, selecting or validating endpoints based
on traversability, target proximity, or precomputed manipulation
feasibility~\cite{liu2024okrobot,yan2025dovsg,liu2025dynamem,wang2025instruction}. Such staged treatment provides limited support
for continuously refining the navigation endpoint as local reachability
and collision conditions change. Consequently, a nearby endpoint may
remain infeasible for manipulation, while execution failures may not
revise the base pose, navigation plan, or scene memory. OmniNav instead
casts interaction-feasible base-pose selection as continuous optimization
over a differentiable dual-ellipsoid approximation of the manipulator
workspace, with collision safety encoded by geometric penalties. It
further propagates execution feedback through hierarchical recovery,
from local operation retry to base-pose re-optimization and memory-guided
target re-search.

The main contributions are summarized as follows:

\begin{itemize}
    \item We present OmniNav, a training-free framework that formulates
long-horizon execution as factorized probabilistic inference over
coupled scene memory, target search, and interaction feasibility states
in dynamic environments.

      \item We develop a dynamically revisable object-centric 3D memory that uses
      Bayesian persistence inference to remove stale evidence while preserving
      structured scene context in a hierarchical scene graph.

      \item We introduce an exploration-conditioned Bayesian target search
      strategy that revises dependency-aware region priors with unsuccessful
      search evidence and verifies fine-grained candidates through dual-source
      confirmation.

      \item We propose a reachability-aware closed-loop manipulation module that combines differentiable dual-ellipsoid-based base-pose optimization with hierarchical recovery.

      \item Extensive evaluations across navigation benchmarks, dynamic long-horizon tasks, and real-world mobile manipulation demonstrate greater robustness and efficiency, raising pick-and-place success from 53.3\% to 71.7\%.
\end{itemize}

\section{Related Work} \label{RW}

We review embodied navigation research from three perspectives:
scene representation and memory, target-oriented navigation, and
interaction-aware mobile manipulation.

\subsection{Scene Representation and Memory}

Scene memory and representation are fundamental to embodied navigation. Early end-to-end navigation methods\cite{maksymets2021thda,ramrakhya2023pirlnav} directly map sensory observations to actions without maintaining explicit scene memory, limiting their interpretability and adaptability in long-horizon real-world environments. Although several recent approaches introduce implicit memory mechanisms\cite{zhang2024uni,zeng2024poliformer}, their latent representations remain difficult to update under evolving scene dynamics. In contrast, modular navigation frameworks based on explicit scene representations offer greater interpretability, modularity,
  and direct support for persistent scene reasoning.

Early modular methods\cite{chaplot2020object,ramakrishnan2022poni} primarily rely on 2D or 3D semantic maps for object-goal navigation within predefined semantic categories. For example, semantic observations can be projected into global occupancy maps or semantic point clouds to guide navigation planning. With the emergence of pretrained vision-language models (VLMs)\cite{radford2021learning,li2023blip2,zhai2023sigmoid}, recent mapping approaches have begun to construct open-vocabulary scene 
representations by projecting visual-language features onto spatial maps. VLMaps\cite{huang2023visual} builds 2D grid maps with region-level CLIP features, while ConceptGraphs\cite{gu2024conceptgraphs} extends this idea to 3D point-cloud-based open-vocabulary scene representations. While these methods offer strong open-vocabulary mapping and querying, our focus is complementary: online mapping under partial observability, instance-level validity revision, and their integration with target-directed exploration.
To improve online navigation efficiency, another line of work\cite{yokoyama2024vlfm,zhang2025apexnav} constructs target-conditioned semantic maps based on vision-language similarity. Although these methods enable efficient online navigation, their scene representations are primarily organized around target relevance rather than persistent object-centric scene understanding, limiting their long-term reusability across diverse embodied tasks.

More recently, scene-graph-based methods~\cite{yin2024sg,yin2025unigoal}
  represent environments through structured object-centric memories that encode
  object concepts and semantic or geometric relations. SG-Nav~\cite{yin2024sg}
  constructs online 3D scene graphs and uses LLM-based relational reasoning for
  zero-shot navigation, while UniGoal~\cite{yin2025unigoal} unifies multiple
  goal modalities through graph-based scene--goal matching. These structured
  representations improve target grounding and relational reasoning, but are
  largely developed under static-scene assumptions and do not explicitly model
  the temporal validity of stored instances.

Change-aware mapping methods model scene evolution at the representation
level~\cite{qian2022pocd,RSS24Khronos}, but do not connect map-validity
updates with target-directed exploration and navigation. Recent embodied navigation systems maintain dynamic scene knowledge through
carrier- or anchor-based updates. OpenIN~\cite{tang2025openin} and OpenObject-NAV\cite{tang2025openobject} revise carried
objects relative to previously mapped carriers, whereas DualMap~\cite{jiang2025dualmap} constructs its map
online but models dynamics primarily as volatile-object reassociations with
assumed-static anchors. OVSE
~\cite{bogenberger2026did} instead revisits potentially stale regions to
refresh semantic evidence. These methods improve navigation under scene
changes, but provide limited mechanisms for inferring the validity of
individual stored instances from partial and ambiguous observations.
OmniNav instead performs instance-level Bayesian persistence inference from supporting,
empty-space, and occlusion evidence to continuously revise the validity
of stored objects.

\subsection{Target-Oriented Navigation Strategies}

Target-oriented navigation strategies determine where a robot should move next during target search. A representative paradigm in modular navigation is frontier-based exploration, which iteratively identifies the boundary between explored and unexplored regions and plans paths toward selected frontiers to expand spatial coverage. For example, CoW~\cite{gadre2023cows} adopts a nearest-frontier strategy for open-vocabulary object search.

To introduce semantic guidance into exploration, recent methods score candidate frontiers according to their relevance to the target. ESC~\cite{zhou2023esc} and L3MVN~\cite{yu2023l3mvn} convert frontier-level observations into text-based representations and leverage LLMs for commonsense reasoning. ESC encodes commonsense knowledge, such as ``\textit{fridges are in kitchens}'', as probabilistic soft logic constraints, while L3MVN describes frontier contents in natural language and uses an LLM to rank promising candidates. SG-Nav~\cite{yin2024sg} additionally uses object--room relations to guide
  candidate selection. Instead of relying primarily on language reasoning, VLFM~\cite{yokoyama2024vlfm} selects frontier points using BLIP-2-based~\cite{li2023blip2} vision-language similarity between observations and the target. Building upon this VLM similarity criterion, ApexNav~\cite{zhang2025apexnav} further combines semantic and geometric information to adaptively guide exploration according to the environment's semantic distribution.

While these methods improve target search within individual navigation episodes, a parallel line of work explores unified strategies that support multiple navigation tasks under a shared decision framework. UniGoal~\cite{yin2025unigoal} unifies object, image, and fine-grained
text-goal navigation through graph-based scene--goal matching. InstructNav~\cite{long2024instructnav} introduces a Dynamic Chain-of-Navigation (DCoN) that standardizes heterogeneous navigation instructions, such as object search, vision-language navigation, and demand-driven navigation, into a unified action--landmark chain, and fuses multi-source value maps for decision-making.

Despite these advances in semantic exploration and unified navigation, region or landmark selection is still driven primarily by positive relevance or value cues. Although conventional
  coverage maps record which areas have been explored, few methods explicitly
  interpret target-absent observations as negative evidence that revises a
  region's target probability. OmniNav formulates this process as
  exploration-conditioned Bayesian belief revision, updating dependency-aware
  semantic priors with accumulated unsuccessful searches.

\subsection{Interaction-Aware Mobile Manipulation}
Mobile manipulation requires the robot to reach an interaction-feasible base
  pose rather than merely approach the target\cite{yokoyama2023asc,lin2026affordance}. A geometrically close pose may still be unsuitable because of manipulator reachability constraints or local collision risks. OK-Robot~\cite{liu2024okrobot} connects
  target navigation with grasping and placing, but follows a sequential
  pipeline based on a one-time scene scan, providing limited adaptation to
  scene changes and execution failures. IALP~\cite{wang2025instruction}
  introduces grounded manipulation-feasibility checking through a pre-built
  scene map and a precomputed robot-centric reachability map, but uses reachability only as a staged feasibility test, rather than for continuous collision-aware base-pose refinement.

  Beyond endpoint feasibility, recent systems also introduce dynamic memory or closed-loop execution.
  DovSG~\cite{yan2025dovsg} maintains a dynamic object-centric scene graph for
  mobile manipulation but relies on a scene map constructed offline. DynaMem
  ~\cite{liu2025dynamem} constructs dynamic spatio-semantic memory online, while
  focusing primarily on representation maintenance rather than manipulation
  reachability and base-pose feasibility. COME-robot~\cite{zhi2025come}
  introduces VLM-based execution verification and failure recovery, but does not
  explicitly optimize the mobile base pose under manipulator-reachability and
  collision constraints. 
OmniNav couples interaction-feasible endpoint optimization with feedback-driven recovery, so physical outcomes revise later base-pose, navigation, and memory decisions.

\begin{figure*}[!t]\centering
	\includegraphics[width=18cm]{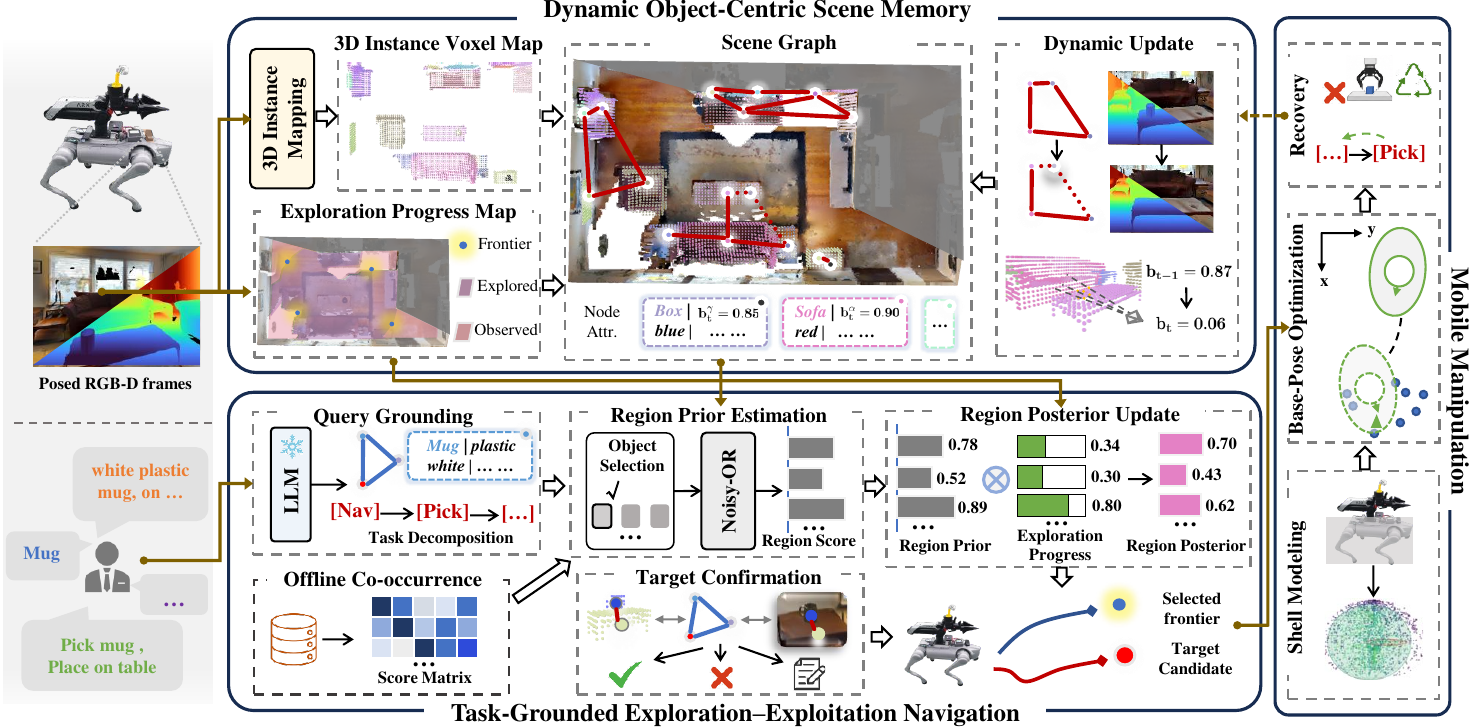}
	\caption{Overall framework of OmniNav. The system consists of three main modules: (i) a dynamic object-centric scene memory that maintains instance-level semantics, geometry, relations, posterior persistence
beliefs $\{b_t^\gamma\}$, and an auxiliary
2D exploration-progress map; (ii) a task-grounded exploration--exploitation navigation strategy that balances semantic target reasoning with exploration progress evidence; and (iii) a reachability-aware closed-loop mobile manipulation module that selects interaction-feasible base poses and recovers from execution failures or scene changes.}
	\label{framework} 
\end{figure*}

\section{Overview} \label{OV}
OmniNav presents a unified framework for long-horizon target navigation and interaction in dynamic environments. At time $t$, the robot receives a posed
RGB-D observation $o_t = (\mathbf{I}_t, \mathbf{D}_t, \mathbf{P}_t)$, where $\mathbf{I}_t$, $\mathbf{D}_t$, and $\mathbf{P}_t$ denote the RGB image, depth image, and
 camera pose, respectively. Given a task query $Q$, the robot must continually revise its understanding of the scene, its judgment of the target,
and the physical feasibility of the resulting navigation decision. No pre-built
map or prior knowledge of the test environment is required.

At the system level, we represent the uncertain long-horizon task
state as
$\mathcal{X}_t=(\mathcal{M}_t,\mathcal{T}_t,\mathcal{F}_t)$,
where $\mathcal{M}_t$, $\mathcal{T}_t$, and $\mathcal{F}_t$
denote the scene memory, target search, and interaction feasibility
states, respectively. Let $\mathcal{Y}_{1:t}$ denote the accumulated
evidence from sensory observations, exploration outcomes, and physical
execution feedback. We treat scene memory inference as query-independent
and let the task query affect interaction feasibility through the
target search state. The task state posterior is then factorized as
\begin{equation}
\begin{aligned}
P(\mathcal{X}_t \mid \mathcal{Y}_{1:t},Q)
&=
P(\mathcal{M}_t,\mathcal{T}_t,\mathcal{F}_t
\mid \mathcal{Y}_{1:t},Q)
\\
&=
P(\mathcal{M}_t \mid \mathcal{Y}_{1:t})
P(\mathcal{T}_t \mid
\mathcal{M}_t,\mathcal{Y}_{1:t},Q)
\\
&\quad \times
P(\mathcal{F}_t \mid
\mathcal{M}_t,\mathcal{T}_t,\mathcal{Y}_{1:t})
\end{aligned}
\label{overall_formulate}
\end{equation}

Sections~\ref{3SM}--\ref{MR} further decompose and evaluate the scene memory,
target search, and interaction feasibility factors, respectively.
Fig. \ref{framework} illustrates the corresponding module-wise instantiation of Eq.~\eqref{overall_formulate}. (i) Dynamic Object-Centric Scene Memory estimates the
scene memory factor by revising the validity of mapped instances from
new observations. (ii) Task-Grounded Exploration--Exploitation
Navigation estimates the target search factor conditioned on the
revised $\mathcal{M}_t$, using semantic context and unsuccessful
exploration to update target belief and identify the intended
candidate. (iii) Reachability-Aware Closed-Loop Mobile Manipulation
instantiates the interaction feasibility factor using the current scene and target states with execution feedback, and refines the final navigation
endpoint according to manipulation reachability and collision safety.
Execution feedback is incorporated into $\mathcal{Y}_{1:t}$ and
triggers re-evaluation of the affected factors.
The major notation used throughout Sections~\ref{3SM}-\ref{MR} is summarized in Table~\ref{tab:notation}.

\begin{table}[!t]
\centering
\caption{Major notation used in OmniNav.}
\label{tab:notation}
\footnotesize
\resizebox{0.95\linewidth}{!}{
\renewcommand{\arraystretch}{1.08}
\setlength{\tabcolsep}{3pt}
\begin{tabularx}{\columnwidth}{
    >{\raggedright\arraybackslash}p{0.32\columnwidth}
    >{\raggedright\arraybackslash}X}
\toprule
\textbf{Symbol} & \textbf{Description} \\
\midrule
$\mathcal{X}_t=(\mathcal{M}_t,\mathcal{T}_t,\mathcal{F}_t)$
&
Coupled scene memory, target search, and interaction feasibility states.
\\
$\mathcal{Y}_{1:t}$
&
Accumulated sensory, search, and execution evidence.
\\
$\Gamma_t,\ z_t^\gamma,\ b_t^\gamma$
& Active instance set, persistence state, and posterior persistence belief of instance $\gamma$. \\

$\mathcal G_t,\ \Gamma_r,\ \Pi^{\mathrm{sp}}$
& Hierarchical scene graph, active instances in region $r$, and supported spatial predicates. \\

$\mathcal K,\ k_{\mathrm{tgt}},\ \mathcal{G}^{Q},\ \mathcal C^{Q}$
& Category vocabulary, target category, query graph, and query constraints. \\

$\xi_r,\ U_r,\ \eta_r$
& Target-existence variable, unsuccessful-search event, and exploration progress of region $r$. \\

$S_{\mathrm{mem}},\ S_{\mathrm{vlm}},\ S_{\mathrm{confirm}}$
& Memory-based, VLM-based, and fused target-confirmation scores. \\


$\psi(\mathbf p),\ \mathcal H,\ \mathbf x$
& Manipulator reachability score, reachability shell, and mobile-base pose. \\
\bottomrule
\end{tabularx}
}
\end{table}

\section{Dynamic Object-Centric Scene Memory} \label{3SM}
This section instantiates the scene memory factor
$P(\mathcal{M}_t\mid\mathcal{Y}_{1:t})$ in
Eq.~\eqref{overall_formulate}. We represent the scene memory state as
$\mathcal{M}_t=(\bar{\mathcal{M}}_t,\mathbf{z}_t,\mathcal{G}_t)$,
where $\bar{\mathcal{M}}_t$ denotes the object-centric map state,
$\mathbf{z}_t$ denotes the joint instance-persistence state, and
$\mathcal{G}_t$ denotes the hierarchical scene graph.
The corresponding posterior factorizes as
\begin{equation}
\begin{aligned}
P(\mathcal{M}_t\mid\mathcal{Y}_{1:t})
={}&
P(\bar{\mathcal{M}}_t\mid\mathcal{Y}_{1:t})
P(\mathbf{z}_t
\mid\bar{\mathcal{M}}_t,\mathcal{Y}_{1:t})
\\
&\times
P(\mathcal{G}_t
\mid\bar{\mathcal{M}}_t,\mathbf{z}_t,\mathcal{Y}_{1:t})
\end{aligned}
\label{eq:memory_decomposition}
\end{equation}

\subsection{Persistent Instance Memory in Dynamic Environments} \label{PIMSSE}

\paragraph{Static Fusion Backbone}
For each incoming RGB frame $\mathbf{I}_t$, OmniNav  extracts instance-level semantics using a two-stage perception pipeline. YOLO-World~\cite{Yolo-world} first detects object instances with open-vocabulary category labels, and TAP~\cite{tap} then produces a pixel-level mask $m_t^k$ and descriptive caption $c_t^k$ for each detection. The captions are encoded by SBERT~\cite{sbert} into semantic embeddings $f_t^k$ for cross-frame association and retrieval:
\begin{equation}
\begin{aligned}
    \label{tap}
    \{m_{t}^{k}, c_{t}^{k}\} &=  \operatorname{SegCap} \bigl( \operatorname{Det}(\mathbf{I}_t) \bigr), \ \
    \{f_{t}^{k}\} = \operatorname{Enc} \bigl( c_{t}^{k} \bigr)
\end{aligned}
\end{equation}

Following OmniMap~\cite{deng2025omnimap}, the instance masks
$\{m_t^k\}$ and semantic embeddings $\{f_t^k\}$ from
Eq.~\eqref{tap}, together with the depth image $D_t$ and camera pose
$P_t$, are projected into 3D and incrementally fused into a
probabilistic voxel map. Each voxel maintains an instance-ID
distribution, while a semantic codebook stores per-instance embeddings.
New detections are associated with existing map instances by
maximum-likelihood assignment over projected voxel sets. At timestep $t$, the resulting map provides the active instance set
$\Gamma_t$, an instance embedding codebook $\mathcal{B}_t$, and an
instance-level voxel representation $\mathcal{V}_t$. We denote the
object-map state obtained from probabilistic observation fusion by
$\bar{\mathcal{M}}_t=(\mathcal{V}_t,\mathcal{B}_t)$. This fusion
instantiates the map factor
$P(\bar{\mathcal{M}}_t\mid\mathcal{Y}_{1:t})$.
We refer readers to OmniMap~\cite{deng2025omnimap} for the static
fusion backbone and focus below on OmniNav's dynamic memory extensions.

\paragraph{Latent Persistence State Model}
While OmniMap provides probabilistic voxel fusion, it lacks mechanisms for detecting and handling object disappearance and relocation, which are essential in dynamic environments. We formulate this problem as online Bayesian inference of instance persistence for each $\gamma\in\Gamma_t$. Each mapped instance is associated with a binary latent persistence state, with a temporal persistence prior and depth-consistency observations from its maintained surface points. To reduce unnecessary computation, OmniNav exploits the asymmetric mobility of indoor objects: movable everyday objects, such as \textit{mugs}, \textit{books}, and \textit{tissues}, are checked whenever observable, whereas large structural objects, such as \textit{tables}, \textit{shelves}, and \textit{cabinets}, are checked at a low fixed frequency.

For each instance $\gamma\in\Gamma_t$, we define a latent binary state $z_t^{\gamma}\in\{0,1\}$, where $z_t^{\gamma}=1$ indicates that the mapped instance remains valid and geometrically consistent at its maintained location, and $z_t^{\gamma}=0$ indicates that the instance has disappeared or been relocated. Collectively, we denote the persistence states of active instances by
$\mathbf{z}_t=\{z_t^\gamma\}_{\gamma\in\Gamma_t}$.
Each persistence state evolves as a two-state Markov chain with an absorbing invalid state:
\begin{equation}
\label{eq:hmm_trans}
\begin{aligned}
P(z_t^{\gamma}=1\mid z_{t-1}^{\gamma}=1) &= 1-\epsilon \\
P(z_t^{\gamma}=0\mid z_{t-1}^{\gamma}=1) &= \epsilon \\
P(z_t^{\gamma}=1\mid z_{t-1}^{\gamma}=0) &= 0 \\
P(z_t^\gamma=0\mid z_{t-1}^\gamma=0) &= 1
\end{aligned}
\end{equation}
where $\epsilon\in(0,1)$ is a small transition prior modeling the
probability that a valid mapped instance becomes invalid between two
consecutive valid persistence-update steps. Once an instance is inferred
invalid at its maintained location, it is not reactivated there; if the
same object is later observed elsewhere, it is instantiated as a new map
instance. 
At the creation time $t_0^{\gamma}$, the initial state prior of $\gamma$ is set to
$P(z_{t_0^{\gamma}}^{\gamma}=1)=1$.

\begin{figure}[t!]\centering
	\includegraphics[width=8.8cm]{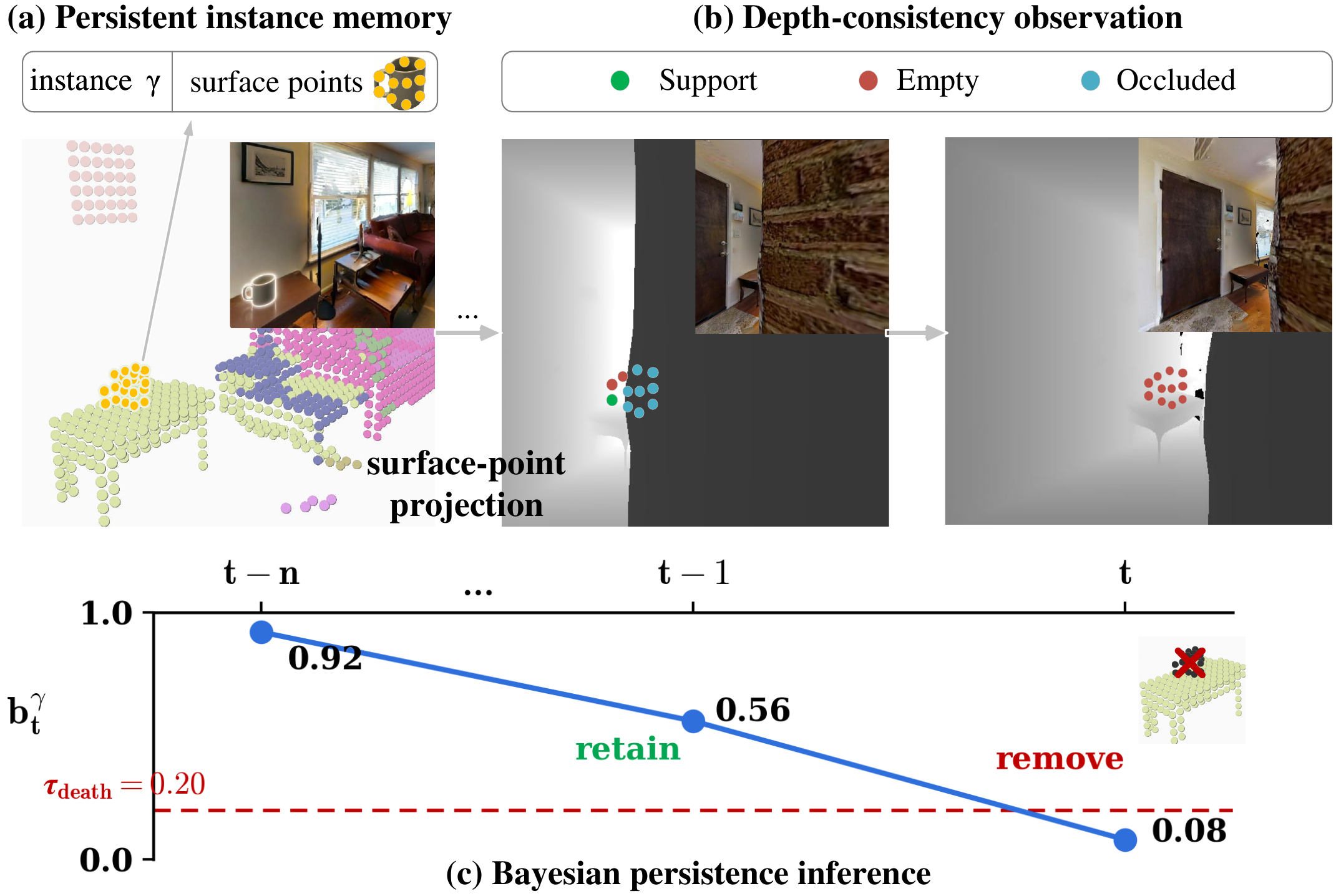}
	\caption{
Illustration of Bayesian persistence inference for dynamic
instance memory.
\textbf{(a)} A remembered instance $\gamma$ is maintained in the instance-level
voxel memory with stored surface points.
\textbf{(b)} The surface points are projected into successive depth observations
and classified as \emph{support}, \emph{empty}, or \emph{occluded}
according to depth consistency. Across $t-n,\ldots,t-1,t$, increasingly
contradictory observations progressively reduce the persistence belief.
\textbf{(c)} The instance is retained for
$b_t^\gamma\geq\tau_{\mathrm{death}}$ and removed from the active
map at its maintained location when
$b_t^\gamma<\tau_{\mathrm{death}}$.
}
	\label{method_dynamic} 
\end{figure}

\paragraph{Surface Point Representation}
To construct the observation model, each instance $\gamma$ maintains a
compact set of surface points
$\mathcal{S}^{\gamma}
=\{\mathbf{s}_i^{\gamma}\in\mathbb{R}^3\}_{i=1}^{N_{\gamma}}$,
sampled from the occupied voxels assigned to the instance in the
object-centric map. Since these voxels are fused from observed depth
surfaces, their world-coordinate positions provide a compact
approximation of the previously observed object surface. Whenever the
instance map is updated, newly fused voxel positions are added to
$\mathcal{S}^{\gamma}$. The set is uniformly subsampled to maintain a bounded representation
size.

\paragraph{Depth-Consistency Observation Likelihood}
At frame $t$, the surface points of instance $\gamma$ are projected into the current camera view using the estimated pose and camera intrinsics. Points outside the image boundaries, behind the camera, or with invalid
depth measurements are discarded. For each remaining valid point $\mathbf{s}_i^\gamma$, let $\hat d_i$ denote its projected depth and $d_i$ denote  the observed depth at the projected pixel. 
To account for range-dependent depth uncertainty, we use an adaptive
tolerance $\tau_d(\hat d_i)=\tau_0+\kappa {\hat d_i}^2$, where
$\tau_0$ is the base tolerance and $\kappa$ controls its
quadratic growth with the projected depth $\hat d_i$. The per-point observation
is categorized as:
\begin{itemize}
    \item \textit{Support}
    $(|d_i-\hat d_i|<\tau_d(\hat d_i))$: the observed depth is consistent
    with the expected object surface;
    \item \textit{Empty}
    $(d_i>\hat d_i+\tau_d(\hat d_i))$: the expected surface location is
    observed as free space;
    \item \textit{Occluded}
    $(d_i<\hat d_i-\tau_d(\hat d_i))$: a nearer surface occludes the
    expected object surface.
\end{itemize}
The frame-level observation is summarized as a count vector 
$\mathbf{n}_t^\gamma=(n_s,n_e,n_o)$ over $N_t^\gamma$ valid projected surface points. 
We model the aggregate counts using a multinomial approximation
conditioned on the latent validity state:
\begin{equation}
\label{eq:obs_model}
\begin{aligned}
\mathbf{n}_t^\gamma \mid z_t^\gamma=1 
&\sim \operatorname{Multinomial}(N_t^\gamma,\boldsymbol{\phi}_1), \\
\mathbf{n}_t^\gamma \mid z_t^\gamma=0 
&\sim \operatorname{Multinomial}(N_t^\gamma,\boldsymbol{\phi}_0)
\end{aligned}
\end{equation}
where $\boldsymbol{\phi}_1=(\phi_s^+,\phi_e^+,\phi_o)$ and
$\boldsymbol{\phi}_0=(\phi_s^-,\phi_e^-,\phi_o)$
are the support, empty-space, and occlusion probability vectors under the
valid and invalid states, respectively.
We assign $\phi_s^-$ a small positive value to account for
accidental depth agreement caused by sensor noise.
$\phi_e^+$ and $\phi_e^-$ model empty-space observations under
valid and invalid states, respectively. A shared occlusion
likelihood $\phi_o$ is used under both states, treating occlusion
as neutral evidence.

\paragraph{Online Bayesian Persistence Inference}
Let $\mathcal{O}_{1:t}^{\gamma}$ denote the depth-consistency observation history used for persistence filtering of instance
$\gamma$ up to frame $t$, where each observation is represented by a support--empty--occluded count vector.
Using these observations as persistence evidence, we approximate the joint persistence posterior with conditionally independent instance filters:
\begin{equation}
\begin{aligned}
P(\mathbf{z}_t
\mid
\bar{\mathcal{M}}_t,\mathcal{Y}_{1:t})
&\approx
P\!\left(
\mathbf{z}_t
\mid
\{\mathcal{O}_{1:t}^{\gamma}\}_{\gamma\in\Gamma_t}
\right)
\\
&=
\prod_{\gamma\in\Gamma_t}
P(z_t^\gamma\mid\mathcal{O}_{1:t}^{\gamma})
\end{aligned}
\label{eq:memory_factorization}
\end{equation}

For each instance, the posterior is recursively evaluated by
\begin{equation}
\begin{aligned}
P(z_t^\gamma\mid\mathcal{O}_{1:t}^{\gamma})
\propto\;&
P(\mathbf{n}_t^\gamma\mid z_t^\gamma)
\\
&\times
\sum_{z_{t-1}^\gamma}
P(z_t^\gamma\mid z_{t-1}^\gamma)
P(z_{t-1}^\gamma\mid
\mathcal{O}_{1:t-1}^{\gamma})
\end{aligned}
\label{eq:persistence_filter}
\end{equation}

We define
$b_t^\gamma=P(z_t^\gamma=1\mid\mathcal{O}_{1:t}^{\gamma})$
as the posterior persistence belief. Marginalizing the transition model
in Eq.~\eqref{eq:hmm_trans} gives the predicted belief
\begin{equation}
\hat b_t^\gamma=(1-\epsilon)b_{t-1}^\gamma
\label{eq:persistence_prediction}
\end{equation}
Combining this prediction with the observation likelihood in
Eq.~\eqref{eq:obs_model} and normalizing gives
\begin{equation}
\label{eq:filter_update}
b_t^{\gamma} =
\frac{P(\mathbf{n}_t^\gamma\mid z_t^\gamma=1)\hat{b}_t^{\gamma}}
     {P(\mathbf{n}_t^\gamma\mid z_t^\gamma=1)\hat{b}_t^{\gamma} + P(\mathbf{n}_t^\gamma\mid z_t^\gamma=0)(1-\hat{b}_t^{\gamma})}
\end{equation}
The prediction and filtering updates are evaluated only when the
instance is scheduled to be checked and sufficient valid
projected surface points are available. Otherwise, the belief remains unchanged,
$b_t^\gamma=b_{t-1}^\gamma$.

An instance is declared invalid at its maintained location when its posterior belief falls below a decision threshold:
\begin{equation}
\label{eq:death_rule}
b_t^{\gamma} < \tau_{\mathrm{death}} \;\Longrightarrow\; 
\Gamma_t\leftarrow\Gamma_t\setminus\{\gamma\}
\end{equation}

\paragraph{Global Map Consistency Maintenance}
When an instance $\gamma$ is declared invalid at its maintained
location via Eq.~\eqref{eq:death_rule}, all voxel entries associated with $\gamma$ are removed from the global map representation to maintain consistency between the voxel memory $\mathcal{V}_t$ and the active instance set $\Gamma_t$. 
This prevents stale object instances from affecting downstream functionality, such as semantic querying.

\subsection{Hierarchical Scene Graph Abstraction} \label{SGSM}
The scene-graph factor in Eq.~\eqref{eq:memory_decomposition} is
instantiated by a lightweight hierarchical scene graph
$\mathcal{G}_t=(\mathcal{N}_t,\mathcal{R}_t,\mathcal{E}_t)$,
conditioned on the current object-map state and inferred persistence
information.
The graph contains instance nodes $\mathcal{N}_t$ for active object instances $\Gamma_t$, region nodes $\mathcal{R}_t$ for spatially grouped scene areas, and relational edges $\mathcal{E}_t$ encoding spatial relations among instances. As the voxel memory is updated, the graph is incrementally synchronized with the active instance set.

\paragraph{Instance Nodes}
Each instance node $n^{\gamma}\in\mathcal{N}_t$ corresponds to an active instance $\gamma\in\Gamma_t$ and serves as the basic object-level entity in the scene graph. The node aggregates information from the voxel memory $\mathcal{V}_t$ and the instance codebook $\mathcal{B}_t$, including semantics, geometry, persistence state, and observation statistics.

\textit{\textbf{Semantics}:} Each node stores a category label $\operatorname{cat}(\gamma)$ initialized from YOLO-World\cite{Yolo-world}, together with the instance
embedding $f^\gamma$ retrieved from the semantic codebook
$\mathcal{B}_t$ for open-vocabulary retrieval. 
To support fine-grained target disambiguation, each node maintains
a structured attribute set $\mathcal{A}^{\gamma}$, including \textit{color}, \textit{material}, \textit{shape}, and \textit{size}. These attributes are extracted from TAP~\cite{tap}
captions using lightweight lexicon matching and aggregated by occurrence counts.

\textit{\textbf{Geometry}:} Each node stores the instance centroid $\mathbf{p}^{\gamma}$ and axis-aligned bounding box $\mathbf{B}^{\gamma}$, which are updated incrementally as new observations refine the instance geometry.

\textit{\textbf{Persistence State}:} Each node maintains the persistence belief $b_t^{\gamma}$ from Eq.~\eqref{eq:filter_update}. 

\textit{\textbf{Observation Statistics}:} 
Each node records observation metadata, including the total number of observations $N_{\mathrm{obs}}^{\gamma}$ and the latest observation timestamp $t_{\mathrm{last}}^{\gamma}$, which provide temporal cues for downstream exploration and target reasoning.

\paragraph{Region Nodes}
To provide a spatial abstraction above individual objects, the scene graph maintains region nodes $\mathcal{R}_t$ that group nearby active instances. 
Each region $r\in\mathcal{R}_t$ contains a subset of active instances $\Gamma_r\subseteq\Gamma_t$, with its centroid computed as the mean of the member-instance centroids. When a new instance is added, it is
assigned to the nearest region if its centroid is within $\tau_{\mathrm{region}}$ of the region centroid and a line segment to at least one member instance is free of intervening occupied geometry. Otherwise, a new region is initialized. Neighboring regions are merged when their centroid distance is below $\tau_{\mathrm{region}}$ and an obstacle-free connection exists between their member instances.
When an instance is removed from $\Gamma_t$, it is also removed from its associated region; empty regions are then deleted.

\paragraph{Relationship Edges}
The edges $\mathcal{E}_t$ encode pairwise spatial relationships between
object instances. Let
$\Pi^{\mathrm{sp}}
=
\{\textit{on},\textit{near},\textit{above},\textit{below}\}$
denote the supported spatial predicate set. Each
predicate is evaluated from geometric attributes, including 3D bounding boxes, relative positions, and overlap.
To avoid repeated online LLM calls, OmniNav queries an LLM
once for each predicate
$\pi\in\Pi^{\mathrm{sp}}$ to synthesize an executable
geometric judgment function
$g_{\pi}:\Gamma_t\times\Gamma_t\rightarrow\{0,1\}$, which is reused 
throughout execution.
For each ordered instance pair
$(\gamma,\gamma')$, the satisfied predicates are stored as:
\begin{equation}
\label{eq:spatial_edge}
e^{\mathrm{sp}}_{\gamma,\gamma'}
=
\left\{
\pi\in\Pi^{\mathrm{sp}}
\;\middle|\;
g_{\pi}(\gamma,\gamma')=1
\right\}
\end{equation}
Natural-language relations in task queries, such as
\textit{``beside''}, are normalized to the corresponding supported
predicates.

After each memory update, the scene graph is synchronized by inserting new instances, removing invalid ones, and re-evaluating affected region assignments and spatial relations.

\subsection{2D Exploration Progress Map}
\label{2D_Exploration_map}

In parallel with the 3D instance-level voxel memory, OmniNav maintains
a 2D exploration-progress map to track spatial coverage during
navigation. The environment is discretized into a 2D grid map, with each cell
assigned one of three states:
$\{\textit{Unknown},\textit{Observed},\textit{Explored}\}$.
\textit{Unknown} cells have not been covered by any valid depth ray,
\textit{Observed} cells have been traversed by at least one such ray but not marked as \textit{Explored},
and \textit{Explored} cells contain the endpoint of a valid depth ray
observed within the central camera field of view.

At each frame, valid depth pixels are back-projected to 3D, transformed
into the world frame using the estimated camera pose, and rasterized
onto the grid. Ray casting promotes cells traversed up to each measured
endpoint from \textit{Unknown} to \textit{Observed}. For measurements
within the central $50^\circ$ horizontal field of view, the endpoint cells are further marked as \textit{Explored}. Cell states are updated monotonically from
\textit{Unknown} to \textit{Observed} and then to \textit{Explored}.

The resulting map records spatial observation progress independently of
the object-centric memory. Its cell states are later aggregated over
scene-graph regions to support exploration--exploitation reasoning.

\section{Task-Grounded Exploration--Exploitation Navigation} \label{EE}
Given the revised scene memory state $\mathcal{M}_t$, this section
instantiates the target search factor
$P(\mathcal{T}_t\mid\mathcal{M}_t,\mathcal{Y}_{1:t},Q)$ in
Eq.~\eqref{overall_formulate}. 
We represent the target search state as
$\mathcal{T}_t=(\boldsymbol{\xi}_t,h_t)$, where
$\boldsymbol{\xi}_t=\{\xi_r\}_{r\in\mathcal{R}_t}$ collects binary
region-level target-existence variables, with $\xi_r=1$ indicating
that region $r$ contains the target, and $h_t$ denotes the target
hypothesis. The target search posterior then factorizes as
\begin{equation}
\begin{aligned}
P(\mathcal{T}_t
\mid\mathcal{M}_t,\mathcal{Y}_{1:t},Q)
={}&
P(\boldsymbol{\xi}_t
\mid\mathcal{M}_t,\mathcal{Y}_{1:t},Q)
\\
&\times
P(h_t
\mid\boldsymbol{\xi}_t,
\mathcal{M}_t,\mathcal{Y}_{1:t},Q)
\end{aligned}
\label{eq:target_decomposition}
\end{equation}

\subsection{Query Grounding}

Given a navigation query $Q$, OmniNav grounds the task description into a structured target specification for scene-graph matching. We consider two types of queries. 
\textbf{Object-goal} queries specify a target category $k_{\mathrm{tgt}}$, such as \textit{``chair''} or \textit{``bottle''}, and are matched to any instance of that category. \textbf{Fine-grained} queries describe a specific target through visual attributes and/or spatial relations, such as \textit{``the white plastic chair next to a table''}, and require extracting explicit attribute and relational constraints to identify the matching instance.

\paragraph{Fine-Grained Instruction Grounding}
For a fine-grained instruction, OmniNav invokes an LLM once at task
initialization to parse the query $Q$ into a structured query
graph
$\mathcal{G}^{Q}
=(\mathcal{N}^{Q},\mathcal{E}^{Q})$.
The graph contains a target node and optional reference-object nodes,
each annotated with a semantic category and a
set of query-relevant visual attributes. In our implementation, the
supported attribute types include \textit{color}, \textit{material}, \textit{shape}, and \textit{size}. 
The spatial relations expressed in the instruction are normalized to the
canonical predicate set $\Pi^{\mathrm{sp}}$ and represented as directed
edges between the corresponding nodes.
We denote the resulting attribute and relational constraints
collectively by $\mathcal{C}^{Q}$. This graph serves as a structured template for candidate matching and requires no further online LLM calls.

\paragraph{Sequential Multi-Target Extension}
For long-horizon tasks involving a sequence of targets 
$Q^{(1)},Q^{(2)},\dots$, OmniNav grounds each query independently when it is activated. Between successive targets, the scene may change as objects are moved or removed. The persistent scene memory in Section~\ref{3SM} tracks observable scene changes, allowing exploration--exploitation decisions for each new target to be
made based on the updated scene state. Invalidated instances are excluded from candidate consideration, while previously built object and region memories are reused for subsequent search. This enables OmniNav to chain multiple navigation goals without rebuilding the scene representation from scratch.
\subsection{Region Prior Estimation}
Before accounting for exploration progress, OmniNav first estimates a target-existence prior over the regions constructed from online observations. This prior is derived from commonsense object co-occurrence knowledge
and observed object context, reflecting region-level
semantic plausibility.

\paragraph{Offline Object Co-occurrence Priors}
To capture commonsense object co-occurrence patterns, we construct an offline object co-occurrence prior from image-level co-occurrence statistics in Visual Genome~\cite{krishna2017visual}. We first define an extensible category set $\mathcal{K}$ by normalizing common indoor object names and aligning them with the open-vocabulary detector labels. For each category pair $(k_i,k_j)$, we estimate
$q(k_i,k_j)=P(k_j\mid k_i)\in[0,1]$ from their co-occurrence frequency across images, reflecting the likelihood that category $k_j$ appears in an image containing category $k_i$. The resulting co-occurrence matrix is computed offline and stored as a reusable knowledge base for region-level prior estimation.

\begin{figure}[t!]\centering
	\includegraphics[width=9cm]{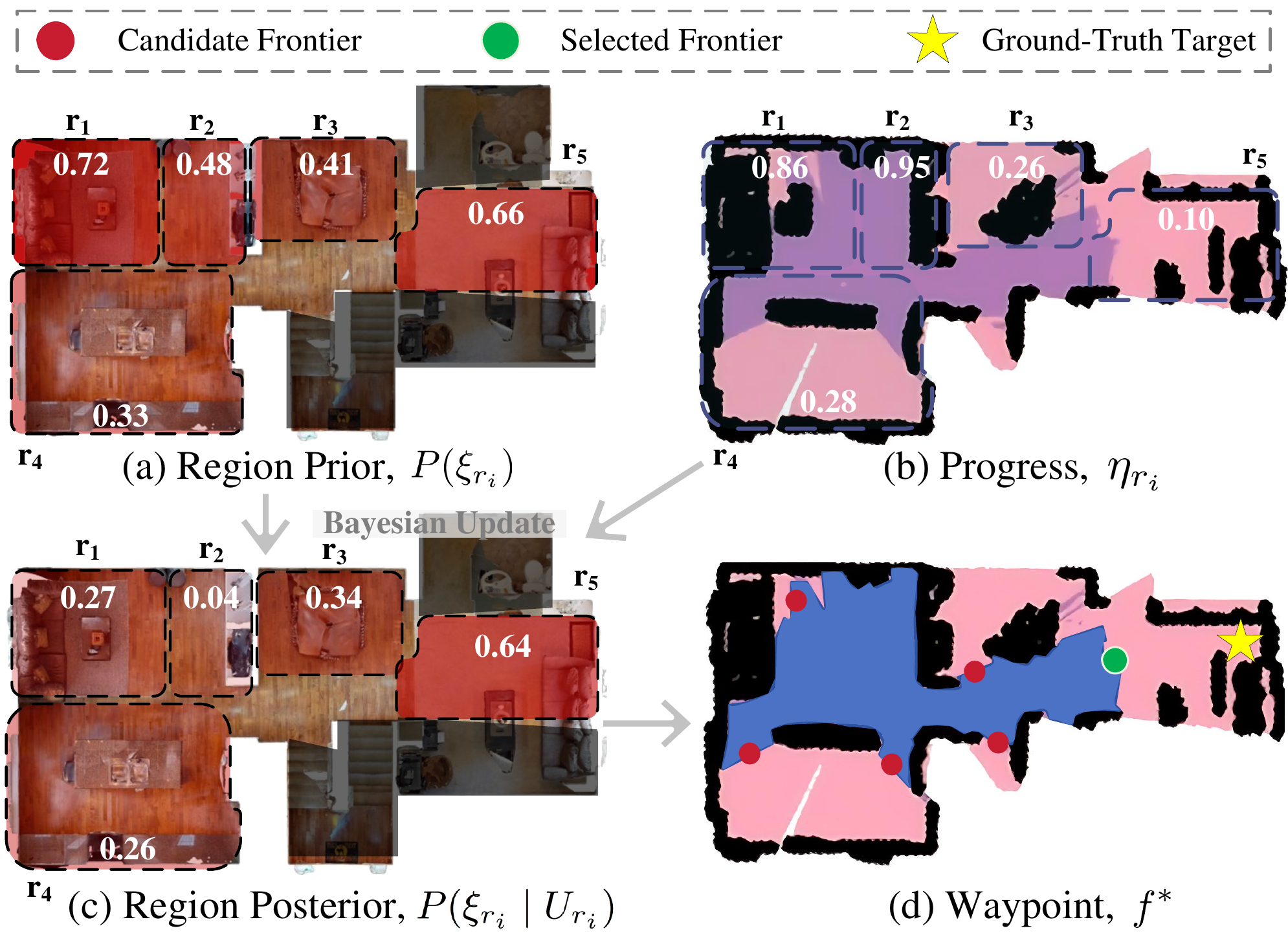}
	\caption{Updating region-level target priors based on exploration progress.
\textbf{(a)} OmniNav first estimates a region-level target prior $P(\xi_{r_i})$ from object co-occurrence cues.
\textbf{(b)} The 2D exploration progress map estimates the exploration progress $\eta_{r_i}$ of each region, where sufficiently explored regions provide negative evidence if the target has not been detected.
\textbf{(c)} The prior is updated into the posterior $P(\xi_{r_i}\mid U_{r_i})$, down-weighting explored-but-unsuccessful regions.
\textbf{(d)} The posterior guides waypoint selection $f^*$ for efficient target search.}
	\label{explo_progress_posterior}
\end{figure}

\paragraph{Region-Level Prior Aggregation}
Let $\bar \xi_r$ denote the event $\xi_r=0$. Given a region $r$ with instances $\Gamma_r=\{\gamma_1,\ldots,\gamma_{N_r}\}$, OmniNav estimates a region-level target prior by aggregating object co-occurrence evidence. 
For each instance $\gamma_i\in\Gamma_r$, its expected target evidence
under the current persistence posterior is
\begin{equation}
\begin{aligned}
a_i
&=
\mathbb{E}\!\left[
z_t^{\gamma_i}
\mid
\mathcal{O}_{1:t}^{\gamma_i}
\right]
q\!\left(
\operatorname{cat}(\gamma_i),k_{\mathrm{tgt}}
\right)
\\
&=
b_t^{\gamma_i}
q\!\left(
\operatorname{cat}(\gamma_i),k_{\mathrm{tgt}}
\right)
\end{aligned}
\label{eq:object_target_evidence}
\end{equation}

When combining these object-level cues,
directly aggregating object co-occurrence scores may overestimate the
target prior, as correlated context objects are treated as independent
evidence. For example, \textit{desks} and
\textit{chairs} often represent the same scene context. We therefore introduce a \textit{Dependency-Aware Region Prior Aggregation} (DA-RPA) strategy, which groups dependent objects and selects representative evidence to compute the region-level target prior.

Specifically, we construct a dependency graph within each region. For two instances $\gamma_i$ and $\gamma_j$, their dependency strength is defined as
\begin{equation}
\rho_{ij}
=
\max
\Bigl\{
q\!\left(\operatorname{cat}(\gamma_i),\operatorname{cat}(\gamma_j)\right),
q\!\left(\operatorname{cat}(\gamma_j),\operatorname{cat}(\gamma_i)\right)
\Bigr\}
\end{equation}

An edge is added between $\gamma_i$ and $\gamma_j$ if $\rho_{ij}>\tau_{\mathrm{dep}}$. Let
$\{\mathcal{D}_m\}_{m=1}^{M_r}$ denote the connected components of the resulting dependency graph in region $r$, where $M_r$ is the number of components. Each component is treated as a correlated evidence group, whose target evidence is computed as
\begin{equation}
\tilde{a}_m = \max_{\gamma_i\in\mathcal{D}_m} a_i
\end{equation}

Treating the resulting dependency groups as approximately independent, the memory-conditioned semantic prior is estimated using the Noisy-OR model\cite{srinivas1993generalization}:
\begin{equation} 
P(\xi_r\mid\mathcal{M}_t,Q)
=
1-\prod_{m=1}^{M_r}
\left(1-\tilde{a}_m\right)
\label{eq:noisy_or}
\end{equation}

\subsection{Exploration-Progress-Aware Posterior Update} \label{ExpProg}
The prior $P(\xi_r\mid\mathcal{M}_t,Q)$ captures semantic co-occurrence cues but does not reflect how thoroughly region $r$ has been searched. We therefore update it using the region-level exploration progress derived from the 2D
exploration progress map.

To associate exploration states with scene-graph regions, grid cells
intersecting the occupied footprint of any instance
$\gamma\in\Gamma_r$ are assigned to region $r$. Each remaining cell
is assigned to the nearest region whose centroid lies within a radius
$r_{\max}$. Cells that cannot be associated with any region are treated
as background and excluded from region-level aggregation.
Let $Cell_r$ denote the set of cells assigned to region $r$. We define $N^{\,r}_{\mathrm{Explored}}$ and $N^{\,r}_{\mathrm{Observed}}$ as the numbers of  cells in $Cell_r$ marked as \textit{Explored} and \textit{Observed}, respectively. The exploration progress of region $r$ is defined as
\begin{equation}
\label{eq:expl_prog}
\eta_r = \frac{N^{\,r}_{\mathrm{Explored}}}{N^{\,r}_{\mathrm{Explored}} + N^{\,r}_{\mathrm{Observed}}} 
\end{equation}

Intuitively, a region becomes less likely to contain the target as
more of its observed area is confidently explored without detecting the target. We model this as negative exploration evidence. 
Let $U_r$ denote the event that the target has not been found in region
$r$ under the current exploration progress $\eta_r$. 
Using $U_r$ as the region-wise summary of accumulated search evidence,
we maintain conditionally independent region beliefs given the current
scene memory and task query:
\begin{equation}
P(\boldsymbol{\xi}_t
\mid\mathcal{M}_t,\mathcal{Y}_{1:t},Q)
=
\prod_{r\in\mathcal{R}_t}
P(\xi_r
\mid U_r,\mathcal{M}_t,Q)
\label{eq:region_factorization}
\end{equation}

For compactness, the conditioning on the current
$\mathcal{M}_t$ and $Q$ is omitted in the following region-level
update. Under a uniform search model, the unsuccessful-search
likelihood is
\begin{equation}
P(U_r\mid\xi_r)=1-\eta_r,
\qquad
P(U_r\mid\bar{\xi}_r)=1
\label{eq:search_likelihood}
\end{equation}
 
Combining this likelihood with the region prior in
Eq.~\eqref{eq:noisy_or}, Bayes' rule gives
\begin{equation}
\label{eq:ee_posterior}
P(\xi_r\mid U_r)
=
\frac{
(1-\eta_r)P(\xi_r)
}{
(1-\eta_r)P(\xi_r)
+
P(\bar \xi_r)
}
\end{equation}

Eq.~\eqref{eq:ee_posterior} satisfies two intuitive boundary conditions under this uniform search model: $P(\xi_r\mid U_r)=P(\xi_r)$ when $\eta_r=0$, and $P(\xi_r\mid U_r)=0$ when $\eta_r=1$. The posterior also decreases monotonically with $\eta_r$. Thus, regions that have been extensively explored without detecting the target receive progressively lower posterior probability, redirecting search toward less-explored regions that remain semantically plausible.  Fig.~\ref{explo_progress_posterior} illustrates how region-level priors are updated to obtain posteriors conditioned on exploration progress.

\vspace{-0.3cm}
\subsection{Navigation Decision} \label{EEDecision}
Given the posterior beliefs over regions, the robot selects the next navigation waypoint from the set of frontier cells. In our
exploration-progress map, frontiers are defined as reachable and
traversable cells lying on the boundary between \textit{Explored} and \textit{Observed} areas.
For each frontier cell $f$, OmniNav computes a utility score that balances target likelihood and navigation efficiency. Let $r(f)$ denote the region associated with frontier $f$, and let $d_f$ be the shortest-path distance from the robot to $f$. We define the normalized distance as
\begin{equation}
\tilde{d}_f = \min(d_f,d_{\max})/d_{\max}
\end{equation}
where $d_{\max}>0$ is the path-distance clipping threshold used for
normalization. The frontier score is then computed as
\begin{equation}
\label{eq:frontier_score}
\operatorname{score}(f)
=
\lambda P(\xi_{r(f)} \mid U_{r(f)})
+
(1-\lambda)(1-\tilde{d}_f)
\end{equation}
where $\lambda\in[0,1]$ balances target-posterior guidance and travel efficiency. For frontiers not associated with any region, the posterior term is assigned a small background prior. The robot selects the highest-scoring frontier:
\begin{equation}
\label{eq:frontier_select}
f^*
=
\arg\max_f \operatorname{score}(f)
\end{equation}
This process is repeated until a target candidate is identified or all reachable frontiers are exhausted. For fine-grained queries, identified candidates are passed to the confirmation stage.

\begin{figure}[t!]\centering
	\includegraphics[width=9cm]{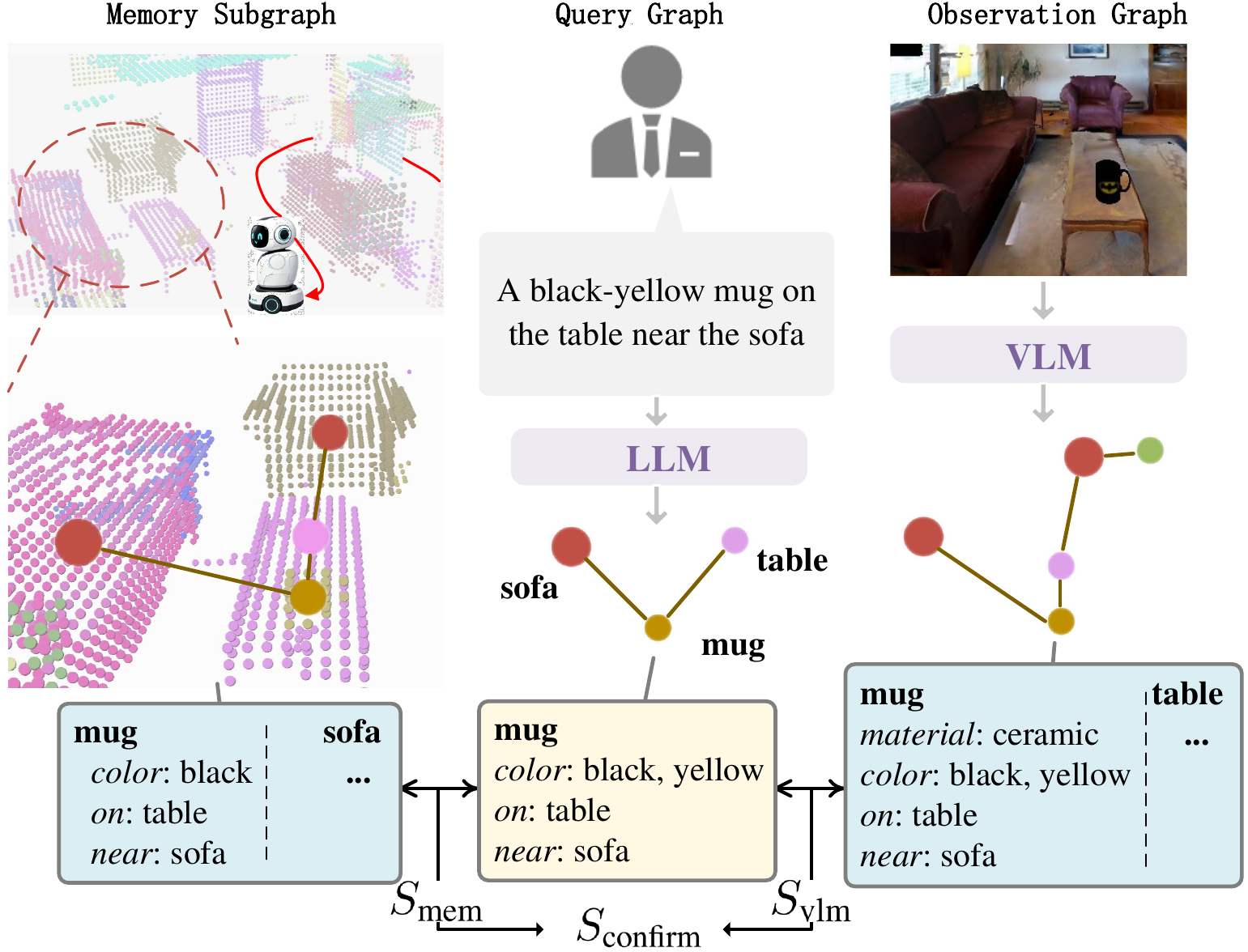}
	\caption{Dual-source fine-grained target confirmation.
The instruction is converted into a query graph $\mathcal{G}^{Q}$
encoding target attributes and relations.
For candidate $\gamma$, matching it with
$\mathcal{G}_t^{\mathrm{mem}}(\gamma)$ and
$\mathcal{G}_t^{\mathrm{vlm}}(\gamma)$ produces
$S_{\mathrm{mem}}$ and $S_{\mathrm{vlm}}$, which are fused into
$S_{\mathrm{confirm}}$. Shared node colors indicate corresponding object roles
across the three subgraphs.}
	\label{method_fine_grained} 
    \vspace{-0.5cm}
\end{figure}
\vspace{-0.5cm}
\subsection{Fine-Grained Confirmation and Exploration--Exploitation Transition} \label{FGTC}

For object-goal queries, the target-hypothesis factor reduces to
category matching, and a category-matched candidate can be accepted
directly. For fine-grained queries, once a target candidate is
identified, OmniNav verifies whether it satisfies the query by matching
the query graph $\mathcal{G}^{Q}$
against two complementary sources: the persistent scene memory and
the current egocentric observation. The memory source provides
accumulated multi-view semantic and relational context, whereas the
egocentric source provides up-to-date fine-grained visual evidence.
The two source-specific matching scores are fused into a confirmation score. 
Let $\Gamma_t^{\mathrm{cand}}\subseteq\Gamma_t$ collect the candidate instances identified during navigation.
An observed instance matching the target category $k_{\mathrm{tgt}}$ is treated as a candidate and added to $\Gamma_t^{\rm cand}$.

\paragraph{Dual-Source Candidate Subgraph Construction}

Given a candidate instance $\gamma$, OmniNav constructs two local subgraphs describing the candidate and surrounding context.

\textit{Memory  subgraph $\mathcal{G}_t^{\mathrm{mem}}(\gamma)$.}
OmniNav extracts from $\mathcal G_t$ a local subgraph centered at the
candidate $n^\gamma$, containing its neighboring instance nodes and
relational edges. Each node $n^{\gamma'}$ retains its attribute set
$\mathcal A^{\gamma'}$, while $n^\gamma$ additionally retains its
observation count $N_{\mathrm{obs}}^\gamma$.

\textit{VLM observation subgraph} $\mathcal{G}_t^{\mathrm{vlm}}(\gamma)$.
In parallel, an egocentric RGB observation containing
the candidate and its surrounding visual context is provided to a VLM.
The VLM is prompted to generate a structured graph using a node-and-edge schema
compatible with the query graph $\mathcal{G}^{Q}$ and the
memory subgraph $\mathcal{G}_t^{\mathrm{mem}}(\gamma)$. The resulting graph $\mathcal{G}_t^{\mathrm{vlm}}(\gamma)$ contains object categories, visual attributes, and pairwise spatial relations inferred from the current viewpoint.


\paragraph{Constraint Matching}





For each information source
$x\in\{\mathrm{mem},\mathrm{vlm}\}$, every query constraint $c\in\mathcal{C}^{Q}$ is
evaluated against the corresponding evidence in the source-specific candidate subgraph
$\mathcal{G}_t^{x}(\gamma)$, yielding a satisfaction score
$\operatorname{sat}_{x}(c)\in[0,1]$.

For an attribute constraint, the queried attribute is compared with the
corresponding attribute of the candidate or contextual instance using an attribute-specific similarity function.
In our implementation, \textit{color} is evaluated using perceptual color similarity; 
\textit{material} and \textit{shape} labels are
compared using WordNet path similarity; and \textit{size} is evaluated according to
ordinal compatibility.

For a relational constraint, the query target is aligned with the
candidate node, while each reference object is matched to a compatible
contextual node. Satisfaction is determined by whether the corresponding
directed edge carries the queried spatial predicate. Before matching,
relation phrases extracted from the task query and the VLM observation subgraph
are normalized to the canonical predicate set $\Pi^{\mathrm{sp}}$.

Constraints unavailable from a given source are assigned a neutral satisfaction score rather than being treated as mismatches. In contrast, evaluable but inconsistent constraints receive low satisfaction scores.

\paragraph{Dual-Source Score Fusion}

For each source $x\in\{\mathrm{mem},\mathrm{vlm}\}$, the matching score
of candidate $\gamma$ is calculated by averaging the satisfaction scores over all query constraints:
\begin{equation}
\label{eq:single_source_score}
S_x=
\frac{1}{|\mathcal C^\mathcal Q|}
\sum_{c\in\mathcal C^\mathcal Q}
\operatorname{sat}_x(c)
\end{equation}

The memory and VLM matching scores are then fused using an
observation-count-dependent weight:
\begin{equation}
\label{eq:score_fusion}
S_{\mathrm{confirm}} = \alpha(N_{\mathrm{obs}}^{\gamma}) \cdot S_{\mathrm{mem}} \;+\; \bigl(1 - \alpha(N_{\mathrm{obs}}^{\gamma})\bigr) \cdot S_{\mathrm{vlm}}
\end{equation}
where
\begin{equation}
\alpha(N_{\mathrm{obs}}^{\gamma}) = 1 - e^{-N_{\mathrm{obs}}^{\gamma} / N_0} 
\end{equation}
Here, $N_0$ is a saturation constant. As the candidate is observed from more frames, the accumulated memory
evidence receives greater weight. For candidates with limited observation history, the egocentric VLM evidence contributes more strongly to fine-grained confirmation. Fig.~\ref{method_fine_grained} illustrates the query--memory and query--observation matching processes and
their fusion.

For fine-grained queries, the fused confirmation score is used to
instantiate the
target-hypothesis factor in Eq.~\eqref{eq:target_decomposition}:
\begin{equation}
P(h_t=\gamma
\mid
\boldsymbol{\xi}_t,
\mathcal{M}_t,\mathcal{Y}_{1:t},Q)
\propto
S_{\mathrm{confirm}}(\gamma),
\ \
\gamma\in\Gamma_t^{\mathrm{cand}}
\label{eq:candidate_factor}
\end{equation}

\paragraph{Confirmation Decision and Exploration--Exploitation Transition} 

The fused score $S_{\mathrm{confirm}}$ measures the confidence that a candidate instance satisfies the fine-grained query.
If $S_{\mathrm{confirm}}$ exceeds a confirmation threshold $\tau_{\mathrm{confirm}}$, the candidate is accepted as the target, and navigation terminates or proceeds to interaction.
Otherwise, the candidate is retained as a hypothesis, after which OmniNav decides whether to continue exploration or exploit the best candidate hypothesis.

Let $S_{\mathrm{best}}$ denote the highest confirmation score among all current candidate hypotheses. We define the current search progress as
\begin{equation}
\eta_{\mathrm{search}}
=
\frac{N_{\mathrm{Explored}}}
{N_{\mathrm{Explored}}+N_{\mathrm{Observed}}}
\end{equation}
where $N_{\mathrm{Explored}}$ and $N_{\mathrm{Observed}}$  denote the numbers of \textit{Explored} and \textit{Observed} cells in the 2D exploration progress map, respectively. A larger $\eta_{\mathrm{search}}$ indicates more thorough exploration of the observed search area.

To decide whether to commit to a candidate, we compute
\begin{equation}
S_{\mathrm{EE}}
=
S_{\mathrm{best}}
-
\beta(1-\eta_{\mathrm{search}})
\end{equation}
where $\beta$ penalizes premature commitment. When $S_{\mathrm{EE}}$ exceeds a transition threshold $\tau_{\mathrm{EE}}$, OmniNav moves from exploration to exploitation and selects the candidate hypothesis with the highest score. Otherwise, it continues exploring to gather additional evidence and update the candidate scores.
\vspace{-0.2cm}
\section{Reachability-aware Closed-Loop Mobile Manipulation} \label{MR}
After a target is confirmed by the target search module, this section instantiates the interaction feasibility factor
$P(\mathcal{F}_t\mid\mathcal{M}_t,\mathcal{T}_t,\mathcal{Y}_{1:t})$ in Eq.~\eqref{overall_formulate}. OmniNav refines the final navigation
endpoint through reachability-aware base-pose optimization, while execution feedback triggers hierarchical recovery in response to execution failures or state changes.
The navigation endpoint must support downstream manipulation rather than geometric proximity alone, since a nearby pose may still violate reachability or collision constraints.

We focus on the canonical pick-and-place setting\cite{liu2024okrobot}, where the robot navigates to a target object, grasps it, navigates to the target receptacle, and places the object there. Formally, a task is represented as
\begin{equation}
\small
\mathsf{Navigate}(A)\!\to\!
\mathsf{Pick}(A)\!\to\!
\mathsf{Navigate}(B)\!\to\!
\mathsf{Place}(A,B)
\end{equation}
where $A$ and $B$ denote the target and receptacle.

To ensure a reliable interaction, OmniNav introduces two components. First, a reachability-aware base alignment strategy refines the final navigation pose by explicitly considering manipulation feasibility and collision safety. Second, a closed-loop recovery mechanism monitors execution outcomes and triggers grasp correction, base-pose re-optimization, or renewed exploration when failures or scene changes occur.

\begin{figure}[t!]\centering
	\includegraphics[width=8.7cm]{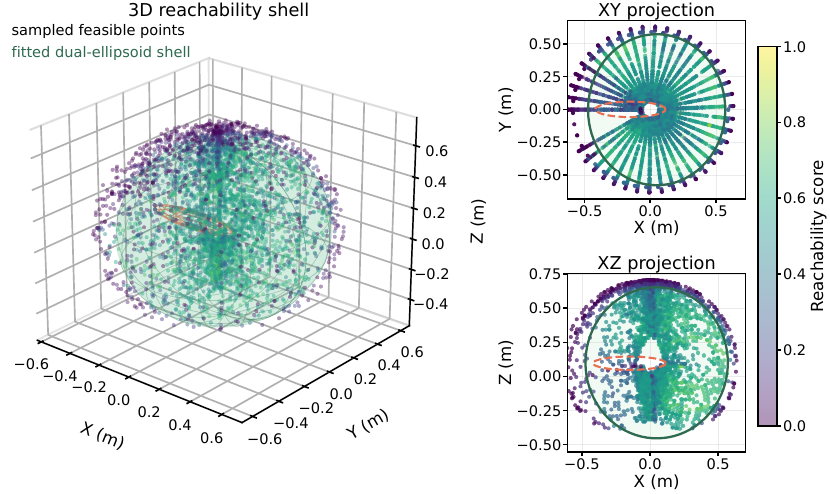}
	\caption{Offline reachability shell modeling. Feasible end-effector positions are sampled through inverse-kinematics-based reachability analysis, and a compact dual-ellipsoid shell is fitted to approximate the high-reachability
workspace. The XY and XZ projections show that the fitted shell captures the main horizontal and vertical reachability structure, providing a smooth geometric surrogate for online base-pose
refinement.}
	\label{dual_ecllips} 
    \vspace{-0.5cm}
\end{figure}

\subsection{Reachability-aware Base Alignment}

\paragraph{Offline Reachability Shell Modeling}

To efficiently evaluate manipulation feasibility during navigation, OmniNav first constructs an offline reachability model of the robot manipulator. 
The model characterizes the region around the mobile base from which targets can be reliably reached, providing an efficient prior for online base alignment.

We sample candidate end-effector positions
$\mathbf{p}\in\mathbb{R}^3$ in the robot base frame. For each sampled position, we construct $N_{\mathrm{IK}}$ end-effector
pose hypotheses by combining $\mathbf{p}$ with uniformly sampled
orientations $\{\mathbf{R}_k\}_{k=1}^{N_{\mathrm{IK}}}$.
In our implementation, $N_{\mathrm{IK}}=20$. A hypothesis is valid if its IK solution satisfies the joint-limit
and self-collision constraints.

The reachability score of position $\mathbf{p}$ is defined as the
fraction of sampled orientations that admit a valid IK solution:
\begin{equation}
\label{eq:point_reachability}
\psi(\mathbf{p})
=
\frac{N_{\mathrm{valid}}(\mathbf{p})}
     {N_{\mathrm{IK}}}
\end{equation}
where $N_{\mathrm{valid}}(\mathbf{p})$ is the
number of orientation hypotheses at $\mathbf{p}$ for which at least one valid IK solution is
found.
A higher $\psi(\mathbf p)$ indicates reachability over a wider range of
end-effector orientations, providing greater manipulation robustness.

We retain positions whose reachability scores exceed a threshold
$\psi_{\mathrm{th}}$:
\begin{equation}
\label{eq:high_reachability}
\mathcal{W}
=
\left\{
\mathbf{p}\in\mathbb{R}^3
\;\middle|\;
\psi(\mathbf{p})\geq\psi_{\mathrm{th}}
\right\}
\end{equation}
The resulting point set $\mathcal{W}$ provides an
orientation-aggregated reachability prior over target positions in the
robot base frame.

Although the sampled reachability set provides a fine-grained characterization of the manipulator workspace, directly querying dense samples is inefficient for online base-pose optimization. We therefore approximate $\mathcal{W}$ with a differentiable dual-ellipsoid reachability shell fitted offline to the retained reachability samples. The outer ellipsoid bounds the reachable workspace:
\begin{equation}
\label{eq:outer_ellipsoid}
d_{\mathrm{out}}(\mathbf{p}) = (\mathbf{p} - \mathbf{c}_{\mathrm{out}})^\top \mathbf{E}_{\mathrm{out}} (\mathbf{p} - \mathbf{c}_{\mathrm{out}})
\end{equation}
while the inner ellipsoid excludes near-base or low-reachability
regions:
\begin{equation}
\label{eq:inner_ellipsoid}
d_{\mathrm{in}}(\mathbf{p}) = (\mathbf{p} - \mathbf{c}_{\mathrm{in}})^\top \mathbf{E}_{\mathrm{in}} (\mathbf{p} - \mathbf{c}_{\mathrm{in}})
\end{equation}
where $\mathbf{p}\in\mathbb{R}^3$ is a query point in the robot base frame, $\mathbf{c}_{\mathrm{out}}$ and $\mathbf{c}_{\mathrm{in}}$ are ellipsoid centers, and $\mathbf{E}_{\mathrm{out}},\mathbf{E}_{\mathrm{in}}\succ 0$ define the ellipsoid shapes. The resulting reachability shell is
\begin{equation}
\label{eq:reachability_shell}
\mathcal{H}
=
\left\{
\mathbf{p}
\;\middle|\;
d_{\mathrm{out}}(\mathbf{p})\le 1,\;
d_{\mathrm{in}}(\mathbf{p})\ge 1
\right\}
\end{equation}

The resulting shell provides a smooth geometric surrogate for manipulation reachability in the subsequent base-pose optimization.
Fig.~\ref{dual_ecllips} illustrates the dual-ellipsoid fitting process and the resulting reachability shell.

\paragraph{Online Base Pose Refinement}
Given the confirmed target hypothesis from $\mathcal{T}_t$, OmniNav
extracts its 3D target position from the current scene state
$\mathcal{M}_t$. For endpoint selection, we represent the
interaction feasibility state by the mobile-base pose
$\mathbf{x}=[x,y,\theta]^\top$ and model its conditional distribution
with an energy function:
\begin{equation}
\begin{gathered}
P(\mathcal{F}_t
\mid
\mathcal{M}_t,\mathcal{T}_t,\mathcal{Y}_{1:t})
=
P(\mathbf{x}
\mid
\mathcal{M}_t,\mathcal{T}_t,\mathcal{Y}_{1:t}),
\\
P(\mathbf{x}
\mid
\mathcal{M}_t,\mathcal{T}_t,\mathcal{Y}_{1:t})
\propto
\exp[-J(\mathbf{x})],
\\
\mathbf{x}^{*}
=
\arg\max_{\mathbf{x}}
P(\mathbf{x}
\mid
\mathcal{M}_t,\mathcal{T}_t,\mathcal{Y}_{1:t})
=
\arg\min_{\mathbf{x}}
J(\mathbf{x})
\end{gathered}
\label{eq:base_pose_distribution}
\end{equation}

The energy $J(\mathbf{x})$ is defined by the composite objective:
\begin{equation}
\label{eq:base_objective}
J(\mathbf{x}) =
w_a J_{\mathrm{align}}(\mathbf{x})
+
w_r J_{\mathrm{reach}}(\mathbf{x})
+
w_c J_{\mathrm{collision}}(\mathbf{x})
\end{equation}
where $w_a$, $w_r$, and $w_c$ are normalized weights satisfying $w_a+w_r+w_c=1$ and control the relative contributions of
target alignment, manipulation reachability, and local obstacle clearance.

The alignment term $J_{\mathrm{align}}(\mathbf{x})$ encourages the robot base to face the target, improving arm accessibility and reducing unnecessary body rotation during execution:
\begin{equation}
\label{eq:align}
J_{\mathrm{align}}(\mathbf{x})
=
1-
\hat{\mathbf{d}}_{\mathrm{base}}(\mathbf{x})^\top
\hat{\mathbf{d}}_{\mathrm{tgt}}(\mathbf{x})
\end{equation}
where $\hat{\mathbf{d}}_{\mathrm{base}}(\mathbf{x})$ is the unit forward direction of the base, and $\hat{\mathbf{d}}_{\mathrm{tgt}}(\mathbf{x})$ is the unit vector from the base position to the target projected onto the ground plane.

To preserve manipulation feasibility, OmniNav considers both target reachability and workspace clearance. Given a candidate base pose $\mathbf{x}$, the target position and local obstacle point cloud extracted from the current map are transformed into the candidate base frame, denoted  $\tilde{\mathbf{p}}_{\gamma}(\mathbf{x})$ and $\tilde{\mathcal{P}}_{\mathrm{obs}}(\mathbf{x})$, respectively. The target reachability penalty is defined as
\begin{equation}
\label{eq:target_reach}
J_{\mathrm{target}}(\mathbf{x})
=
\bigl(d_{\mathrm{out}}(\tilde{\mathbf{p}}_{\gamma}(\mathbf{x}))-1\bigr)_+^2
+
\bigl(1-d_{\mathrm{in}}(\tilde{\mathbf{p}}_{\gamma}(\mathbf{x}))\bigr)_+^2 
\end{equation}
where $(\cdot)_+$ denotes the positive-part operator. This term encourages the target position to lie inside the operable reachability shell.

To avoid obstacle interference with arm motion, OmniNav further penalizes obstacles that intrude into the shell:
\begin{equation}
\label{eq:shell_risk}
J_{\mathrm{shell}}(\mathbf{x})
=
\frac{1}{|\tilde{\mathcal{P}}_{\mathrm{obs}}(\mathbf{x})|}
\sum_{\tilde{\mathbf{p}}\in\tilde{\mathcal{P}}_{\mathrm{obs}}(\mathbf{x})}
\bigl(1-d_{\mathrm{out}}(\tilde{\mathbf{p}})\bigr)_+
\cdot
\sigma\bigl(d_{\mathrm{in}}(\tilde{\mathbf{p}})-1\bigr)
\end{equation}
where $\sigma(\cdot)$ is the sigmoid function. The reachability cost is
\begin{equation}
\label{eq:reach}
J_{\mathrm{reach}}(\mathbf{x})
=
J_{\mathrm{target}}(\mathbf{x})
+
\lambda_{\mathrm{shell}}J_{\mathrm{shell}}(\mathbf{x})
\end{equation}
where $\lambda_{\mathrm{shell}}$ controls the influence of workspace clearance.

A collision penalty is additionally imposed to maintain safe clearance between the robot base and surrounding obstacles. Since obstacle points are evaluated in the candidate base frame, the base center is the origin:
\begin{equation}
\label{eq:chassis}
J_{\mathrm{collision}}(\mathbf{x})
=
\frac{1}{|\tilde{\mathcal{P}}_{\mathrm{obs}}(\mathbf{x})|}
\sum_{\tilde{\mathbf{p}}\in\tilde{\mathcal{P}}_{\mathrm{obs}}(\mathbf{x})}
\zeta\bigl((r_b+\delta_{\mathrm{safe}})-\|\tilde{\mathbf{p}}_{xy}\|\bigr)
\end{equation}
where $r_b$ denotes the chassis radius, $\delta_{\mathrm{safe}}$ is a safety margin, $\tilde{\mathbf{p}}_{xy}$ is the horizontal projection of obstacle point $\tilde{\mathbf{p}}$, and $\zeta(\cdot)$ is the softplus function.

The resulting objective is optimized online using two-stage
coarse-to-fine particle swarm optimization (PSO).
The optimized base pose serves as the final navigation goal, allowing the robot to arrive at an interaction-feasible configuration rather than merely reaching the vicinity of the target.

\subsection{Robust Hierarchical Closed-Loop Recovery}
After reachability-aware base alignment, OmniNav executes manipulation in a closed loop. Failures during long-horizon execution may arise from grasp errors, infeasible base-arm configurations, object displacement, or scene changes. Instead of restarting the task after each failure, OmniNav uses minimal hierarchical intervention, resolving each anomaly at the lowest level that restores execution.

\paragraph{Manipulation-Level Recovery}
This level handles local failures such as grasp slippage or failed grasp closure. 
Using gripper-state and wrist-camera feedback, OmniNav detects the failure, performs local re-perception, updates the grasp or placement pose, and retries without restarting the global task.

\paragraph{Reachability-Level Recovery}
This level is triggered when the target remains identified but the base-arm configuration becomes unsuitable, e.g., due to local target displacement or IK failure. OmniNav re-evaluates geometry, re-optimizes the base pose using the reachability-aware objective, navigates to the interaction-feasible pose, and resumes manipulation.

\paragraph{Task-Level Recovery}
This level is invoked when the task state is invalidated, such as when the target disappears, the confirmed candidate no longer
matches the query, or the receptacle state changes. OmniNav updates the dynamic scene memory, candidate hypotheses and task-relevant object states, and re-enters navigation, target confirmation, or task replanning. 
Execution terminates when the recovery process cannot identify a feasible candidate, reachable pose, or valid task plan.

Execution feedback directly re-evaluates the interaction feasibility
state; task-level failures additionally revise $\mathcal{M}_t$, whose
update subsequently propagates to $\mathcal{T}_t$ and
$\mathcal{F}_t$.

\vspace{-0.2cm}
\section{Simulation Experiments} \label{sim-Exp}
\subsection{Experimental Setup}
We evaluate OmniNav in simulation across three target navigation settings: semantic object navigation\cite{ramakrishnan2021habitat,chang2017matterport3d}, fine-grained instance  navigation\cite{ramakrishnan2021habitat}, and long-horizon object navigation in dynamic environments. These settings are designed to assess different capabilities of OmniNav, including category-level target search, instance-level disambiguation, and adaptation to scene changes. We also conduct systematic ablation studies to evaluate the contribution of each major component, including posterior-guided exploration, fine-grained target confirmation, and dynamic scene memory. Unless otherwise specified, GPT-4o~\cite{hurst2024gpt} is used for both LLM- and VLM-based inference. Our implementation is built primarily with PyTorch, and all simulation experiments are conducted on a single NVIDIA GeForce RTX 4090 GPU.

\begin{table}[!t]
\centering
\caption{Comparison with state-of-the-art methods on  semantic object navigation tasks in HM3D and MP3D.}
\label{ON-table}
\renewcommand\arraystretch{1.15}
\resizebox{0.95\linewidth}{!}{
\setlength{\tabcolsep}{1.5mm}
\begin{tabular}{lccccc}
\toprule
\multirow{2}{*}{\textbf{Method}} & \multirow{2}{*}{\textbf{Training Free}} & \multicolumn{2}{c}{\textbf{MP3D}} & \multicolumn{2}{c}{\textbf{HM3D}} \\
\cmidrule(lr){3-4} \cmidrule(lr){5-6}
 &  & \textbf{SR}$\uparrow$ & \textbf{SPL}$\uparrow$ & \textbf{SR}$\uparrow$ & \textbf{SPL}$\uparrow$ \\
\midrule
SemExp~\cite{chaplot2020object} & $\times$ & 0.360 & 0.144 & -- & -- \\
ZSON~\cite{majumdar2022zson} & $\times$ & 0.153 & 0.048 & 0.255 & 0.126 \\
PONI~\cite{ramakrishnan2022poni} & $\times$ & 0.318 & 0.121 & -- & -- \\
PixNav~\cite{cai2024bridging} & $\times$ & -- & -- & 0.379 & 0.205 \\
SPNet~\cite{zhao2023semantic} & $\times$ & 0.163 & 0.048 & 0.312 & 0.101 \\
\midrule
CoW~\cite{gadre2023cows} & $\checkmark$ & 0.074 & 0.037 & -- & -- \\
ESC~\cite{zhou2023esc} & $\checkmark$ & 0.287 & 0.142 & 0.392 & 0.223 \\
VoroNav~\cite{wu2024voronav} & $\checkmark$ & -- & -- & 0.420 & \cellcolor[HTML]{FFD8B2}0.260 \\
L3MVN~\cite{yu2023l3mvn} & $\checkmark$ & -- & -- & 0.504 & 0.231 \\
InstructNav~\cite{long2024instructnav} & $\checkmark$ & \cellcolor[HTML]{FFD8B2}0.420 & \cellcolor[HTML]{FFFFB2}0.161 & 0.510 & 0.187 \\
GAMap~\cite{yuan2024gamap} & $\checkmark$ & -- & -- & 0.531 & \cellcolor[HTML]{FFD8B2}0.260 \\
SG-Nav~\cite{yin2024sg} & $\checkmark$ & 0.402 & 0.160 & \cellcolor[HTML]{FFFFB2}0.540 & 0.249 \\
UniGoal~\cite{yin2025unigoal} & $\checkmark$ & \cellcolor[HTML]{FFFFB2} 0.410 & \cellcolor[HTML]{FFD8B2}0.164 & \cellcolor[HTML]{FFD8B2}0.545 & \cellcolor[HTML]{FFFFB2}0.251 \\
\midrule
\textbf{OmniNav} & $\checkmark$ & \cellcolor[HTML]{FFB2B2} 0.473 & \cellcolor[HTML]{FFB2B2} 0.180 & \cellcolor[HTML]{FFB2B2}0.546 & \cellcolor[HTML]{FFB2B2}0.285 \\
\bottomrule
\end{tabular}
}
\end{table}

\begin{figure*}[t!]\centering
	\includegraphics[width=17cm]{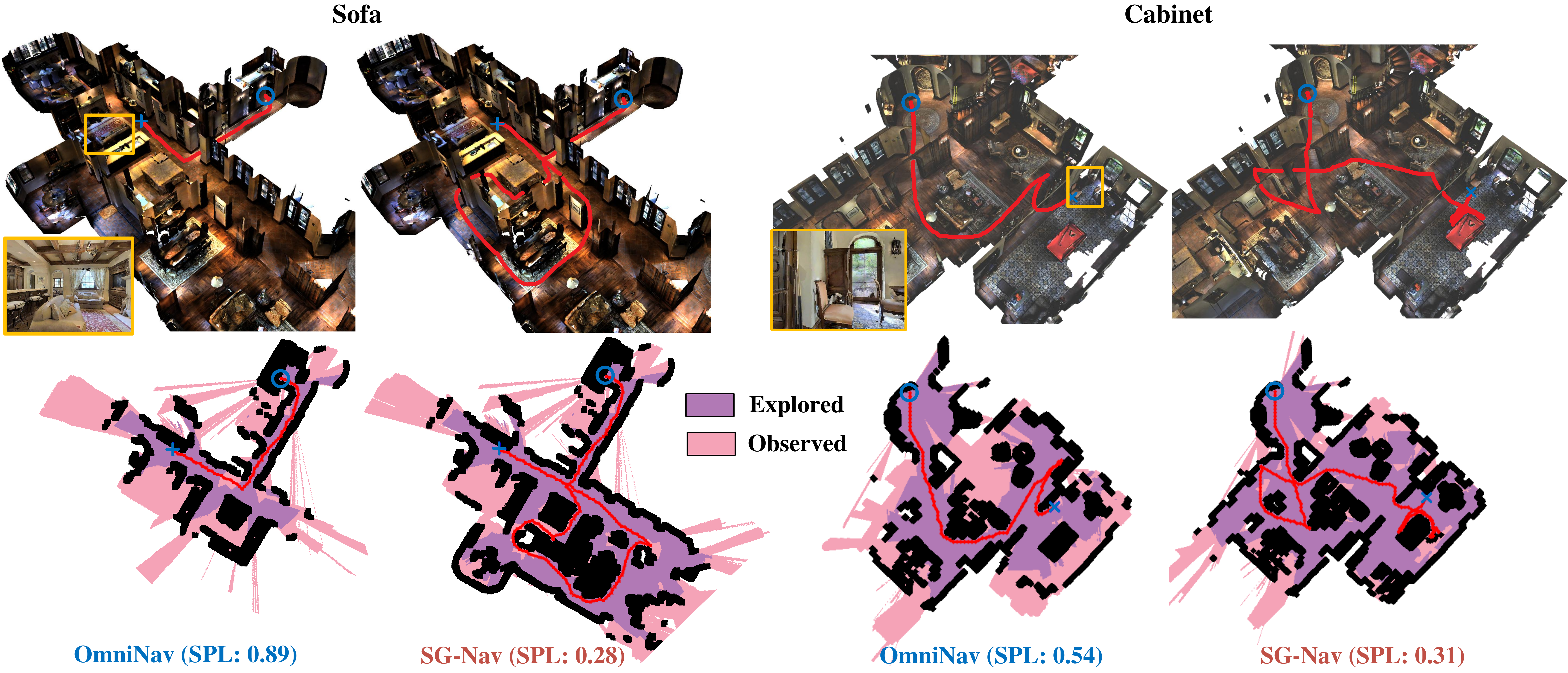}
	\caption{
Qualitative comparison on ObjectNav in MP3D.
Compared with SG-Nav~\cite{yin2024sg}, OmniNav jointly uses semantic co-occurrence and exploration progress information to estimate region-level target posteriors, reducing redundant search in explored-but-unsuccessful regions.
This leads to more efficient trajectories and higher SPL scores: 0.89 vs. 0.28 for \textit{sofa}, and 0.54 vs. 0.31 for \textit{cabinet}.
}
	\label{ex-on}
    \vspace{-0.3cm}
\end{figure*}

\subsection{Semantic Object Navigation}
\label{ON-ex}

We conduct semantic object navigation (ObjectNav) experiments in Habitat on the HM3D\cite{ramakrishnan2021habitat} dataset (2000 episodes, 20 scenes, and 6 goal categories) and the MP3D\cite{chang2017matterport3d} dataset (2195 episodes, 11 scenes, and 21 goal categories).

\textbf{Baselines:} We compare OmniNav with representative semantic object navigation methods, grouped by whether additional task-specific training is required.
Training-based methods include SemExp~\cite{chaplot2020object}, ZSON~\cite{majumdar2022zson}, PONI~\cite{ramakrishnan2022poni}, PixNav~\cite{cai2024bridging}, and SPNet~\cite{zhao2023semantic}. Training-free methods include CoW~\cite{gadre2023cows}, ESC~\cite{zhou2023esc}, VoroNav~\cite{wu2024voronav}, L3MVN~\cite{yu2023l3mvn}, InstructNav~\cite{long2024instructnav}, GAMap~\cite{yuan2024gamap}, SG-Nav~\cite{yin2024sg}, and UniGoal~\cite{yin2025unigoal}.

\textbf{Metrics:} We report Success Rate (SR) and Success weighted by Path Length (SPL), measuring navigation success and efficiency, respectively; higher values are better.

\textbf{Quantitative results:} Table~\ref{ON-table} summarizes the ObjectNav results in the Habitat
simulator. OmniNav achieves the best overall performance among the
compared methods on both MP3D and HM3D. On MP3D, OmniNav improves
the best baseline SR from 0.420 to 0.473 and SPL from 0.164 to 0.180.
On HM3D, it achieves 0.546 SR and 0.285 SPL, compared with the best
baseline values of 0.545 and 0.260, respectively.
By explicitly estimating region-level exploration progress, OmniNav
down-weights previously searched but unsuccessful regions, reducing
redundant revisits and enabling a more effective exploration--exploitation
trade-off. This benefit is reflected in both SR and SPL on MP3D, while
on HM3D it appears more prominently in path efficiency. Even when
successful navigation is achievable, suppressing exhausted regions helps
avoid unnecessary revisits and contributes to the observed SPL improvement.

\begin{table}[t]
\centering
\caption{Ablation study on semantic object navigation tasks in MP3D.}
\label{tab:mp3d_ablation}
\resizebox{0.65\linewidth}{!}{
\begin{tabular}{lcc}
\toprule
& \multicolumn{2}{c}{\textbf{MP3D}} \\
\cmidrule(lr){2-3}
\textbf{Method} & \textbf{SR$\uparrow$} & \textbf{SPL$\uparrow$} \\
\midrule
w/o Exploration Progress & \cellcolor[HTML]{FFFFB2}0.412 & \cellcolor[HTML]{FFFFB2}0.154 \\
w/o DA-RPA & \cellcolor[HTML]{FFD8B2}0.426 & \cellcolor[HTML]{FFD8B2}0.155 \\
\textbf{OmniNav} & \cellcolor[HTML]{FFB2B2}0.473 & \cellcolor[HTML]{FFB2B2}0.180 \\
\bottomrule
\end{tabular}
}
\vspace{-0.3cm}
\end{table}

\begin{figure*}[!t]\centering
	\includegraphics[width=17.5cm]{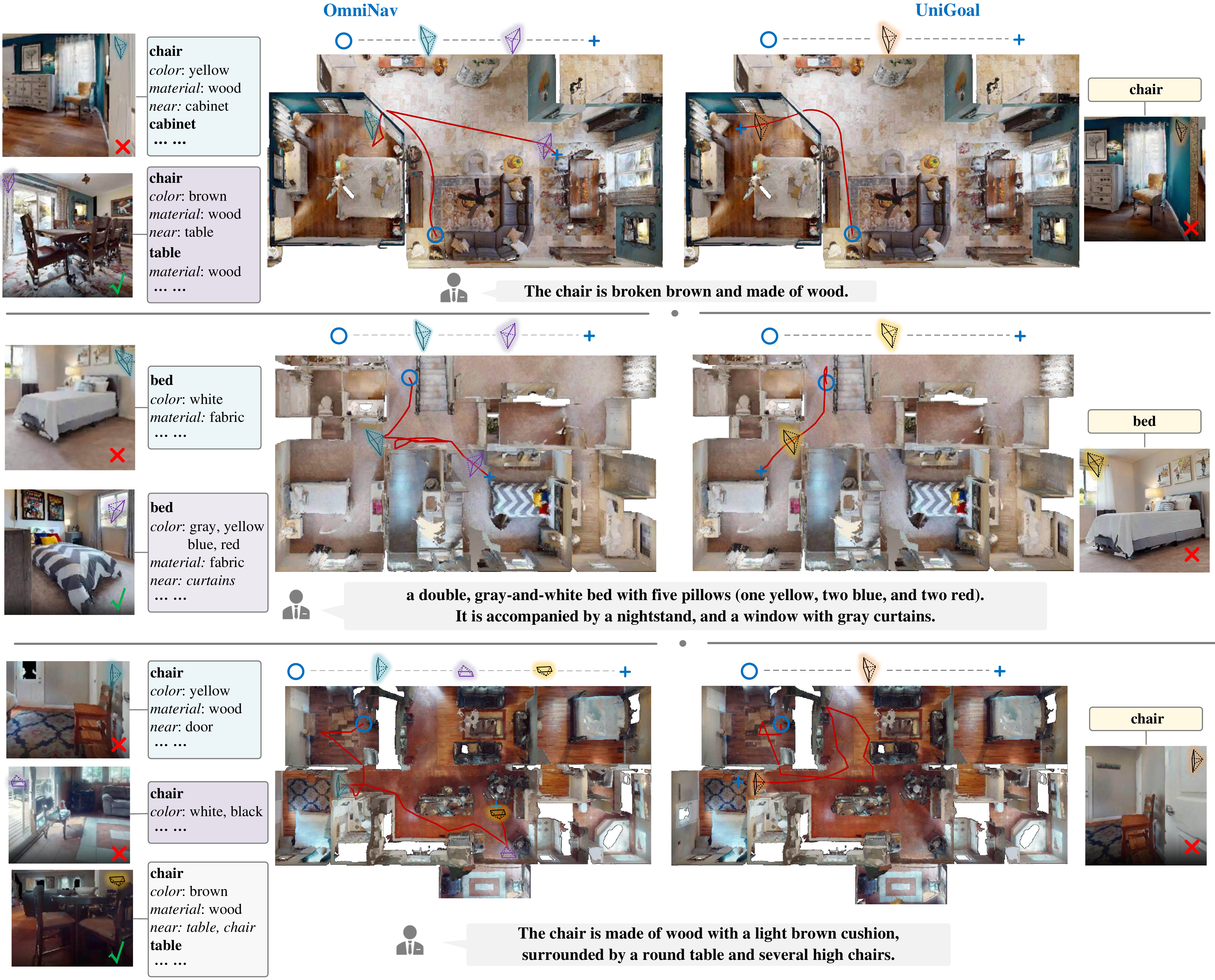}
	\caption{
Qualitative comparison with UniGoal~\cite{yin2025unigoal} on fine-grained instance navigation.
Given instance-specific queries such as \textit{the brown wood chair near the table}, UniGoal may stop at a semantically similar but incorrect \textit{chair} instance.
In contrast, OmniNav verifies both attribute and relational constraints, allowing it to distinguish the target from distractors and complete the task. For clarity, only the VLM-derived observation subgraph is visualized
for each scene.
}
	\label{ex-fine} 
\end{figure*}

\textbf{Qualitative results:}
Fig.~\ref{ex-on} shows qualitative examples on MP3D. Compared with SG-Nav~\cite{yin2024sg}, which primarily relies on semantic relevance for region selection, OmniNav updates region-level target beliefs using both semantic co-occurrence and exploration progress. In the \textit{sofa} and \textit{cabinet} examples, semantic relevance alone can repeatedly attract the robot to locally plausible but already sufficiently searched regions, resulting in redundant exploration and longer trajectories. OmniNav instead down-weights regions where the target has not been found with high exploration progress, and redirects search toward under-explored regions with higher remaining target likelihood. This leads to more efficient trajectories and higher SPL scores, improving from 0.28 to 0.89 in the \textit{sofa} case and from 0.31 to 0.54 in the \textit{cabinet} case.

\textbf{Ablation study:}  We conduct ablation studies on MP3D~\cite{chang2017matterport3d} to evaluate exploration progress modeling and Dependency-Aware Region Prior Aggregation (DA-RPA). As shown in Table~\ref{tab:mp3d_ablation}, removing either component leads to consistent drops in both SR and SPL, while OmniNav achieves the best overall performance. Without exploration progress modeling, OmniNav cannot use unsuccessful search as negative evidence, leading to redundant revisits to semantically plausible but already explored regions. Without DA-RPA, object-level co-occurrence cues are aggregated directly
without dependency grouping, which can overestimate target likelihood
when correlated context objects co-occur. These results validate the contribution of both components to efficient target search.

\begin{table}[t]
\centering
\caption{Comparison with state-of-the-art methods on fine-grained instance-goal navigation in the HM3D dataset.
\textbf{Ins.} indicates whether the method explicitly leverages instance attributes and inter-instance relations for target localization.}
\label{TN-table}
\renewcommand\arraystretch{1.15}
\resizebox{0.8\linewidth}{!}{
\setlength{\tabcolsep}{1.5mm}
\begin{tabular}{lcccc}
\toprule
\multirow{2}{*}{\textbf{Method}} & \multirow{2}{*}{\textbf{Training Free}} & \multirow{2}{*}{\textbf{Ins}} & \multicolumn{2}{c}{\textbf{HM3D}} \\
\cmidrule(lr){4-5} 
 &  & & \textbf{SR}$\uparrow$ & \textbf{SPL}$\uparrow$ \\
\midrule
ZSON~\cite{majumdar2022zson} & $\times$ & $\times$  & 0.106 & 0.049 \\
PSL~\cite{sun2024prioritized} & $\times$ & $\times$  & 0.165 & 0.075 \\
GOAT~\cite{chang2023goat} & $\times$ & $\times$ & 0.170 & 0.088 \\
\midrule
VLFM~\cite{yokoyama2024vlfm} & $\checkmark$ & $\times$ & 0.149 & 0.093 \\
UniGoal~\cite{yin2025unigoal} & $\checkmark$ & $\checkmark$ & \cellcolor[HTML]{FFFFB2}0.202 & \cellcolor[HTML]{FFD8B2}0.114 \\
UniGoal*~\cite{yin2025unigoal} & $\checkmark$ & $\checkmark$ & 0.188 & 0.092 \\
GLMap~\cite{zhang2026multi} & $\checkmark$ & $\checkmark$ & \cellcolor[HTML]{FFD8B2}0.225 & \cellcolor[HTML]{FFB2B2}0.137 \\
\midrule
\textbf{\textbf{OmniNav}} & $\checkmark$ & $\checkmark$ & \cellcolor[HTML]{FFB2B2}0.271 & \cellcolor[HTML]{FFFFB2}0.103 \\
\bottomrule
\end{tabular}
}
\begin{flushleft}
\footnotesize \textit{\textbf{UniGoal*} denotes results reproduced by us using the released open-source implementation of UniGoal~\cite{yin2025unigoal}.}
\end{flushleft}
\vspace{-0.5cm}
\end{table}
\vspace{-0.2cm}
\subsection{Fine-grained Instance Navigation}
We further evaluate fine-grained instance navigation on HM3D (1000 episodes across 36 scenes and 6 target categories). Here, each target is specified by a long-form description of a specific object instance. Each description combines target attributes, such as color and material, with relational context from surrounding objects. 
Accordingly, the benchmark evaluates fine-grained language grounding, relational reasoning, and instance-level discrimination.
We use the same evaluation metrics as in Subsection~\ref{ON-ex}. We compare OmniNav with representative baselines, including training-based methods ZSON~\cite{majumdar2022zson}, PSL~\cite{sun2024prioritized}, and GOAT~\cite{chang2023goat}, as well as training-free methods VLFM~\cite{yokoyama2024vlfm}, UniGoal~\cite{yin2025unigoal}, and GLMap~\cite{zhang2026multi}.

\begin{figure*}[h]\centering
	\includegraphics[width=17.8cm]{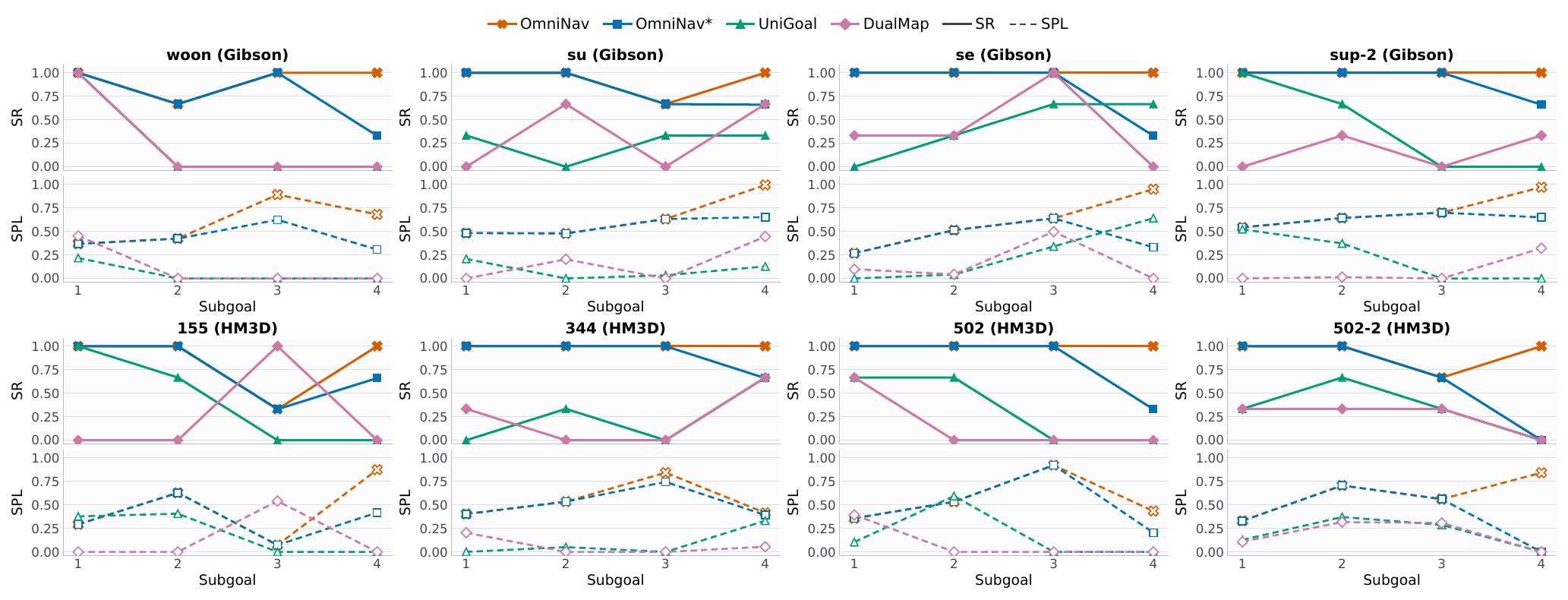}
	\caption{Quantitative results for dynamic long-horizon object navigation in Gibson and HM3D. OmniNav is compared with UniGoal, DualMap, and OmniNav*, a variant without dynamic memory updating, to evaluate scene memory revision during sequential navigation. 
}
	\label{dynamic-ex} 
\end{figure*}

\textbf{Quantitative results:} Table~\ref{TN-table} reports fine-grained instance navigation results. 
OmniNav achieves the highest SR of 0.271, exceeding GLMap\cite{zhang2026multi} and UniGoal\cite{yin2025unigoal}
by 4.6 and 6.9 percentage points, respectively. This improvement is consistent with the benefit of explicit attribute and relational reasoning for instance-level target identification.
Although OmniNav achieves a lower SPL than the strongest baselines, fine-grained navigation requires verifying whether the current candidate matches the instance  description rather than simply reaching the nearest category-level object. 
The confirmation and exploration steps may increase path length, while the higher SR is consistent with improved
target-selection accuracy.

\textbf{Qualitative results}: Fig.~\ref{ex-fine} qualitatively compares OmniNav with UniGoal~\cite{yin2025unigoal} on fine-grained instance navigation. Among the examples, the third task requires finding  \textit{the brown wood chair near the table}, where multiple chair instances with similar categories appear in the scene. UniGoal reaches a semantically relevant region but stops at a same-category distractor. In contrast, OmniNav verifies the candidate using both accumulated scene-graph memory and current egocentric visual evidence, checking the target attributes and its spatial relation to nearby objects. This enables OmniNav to distinguish the correct instance from visually similar distractors and complete the task successfully. The comparison shows that fine-grained navigation benefits from explicit attribute--relation matching and target confirmation before termination.

\begin{table}[!t]
\centering
\caption{Ablation study on fine-grained instance-goal navigation in the HM3D dataset.}
\label{tab:fine_ablation}
\resizebox{0.7\linewidth}{!}{
\begin{tabular}{lc}
\toprule
\textbf{Method} & \textbf{SR$\uparrow$} \\
\midrule
Llama3.2-Vision Backend & \cellcolor[HTML]{FFFFB2}0.256  \\
w/o E-E Transition & \cellcolor[HTML]{FFD8B2}0.265  \\
w/o Egocentric VLM Evidence & 0.231  \\
\textbf{OmniNav} & \cellcolor[HTML]{FFB2B2}0.271   \\
\bottomrule
\end{tabular}
}
\end{table}

\begin{figure*}[t!]\centering
	\includegraphics[width=16cm]{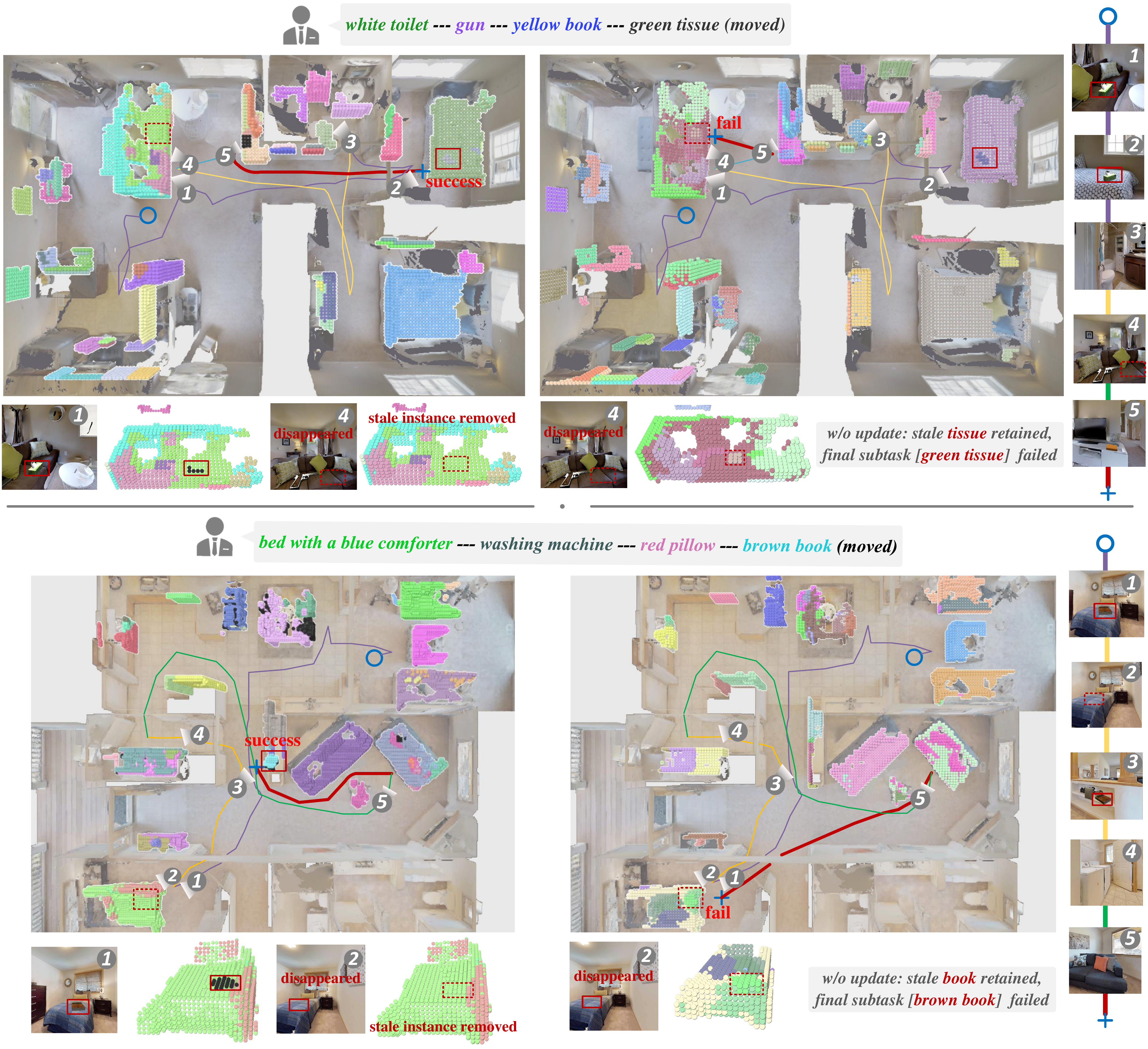}
	\caption{Qualitative comparison on dynamic long-horizon object navigation.
Each episode consists of sequential object-navigation subtasks, where the final target is relocated during execution.
OmniNav (left) detects the scene change, removes stale object instances such as \textit{green tissue} or \textit{brown book}, and reaches the valid final target.
In contrast, OmniNav* without dynamic memory updating (right) retains the stale instance at its original map location, causing the robot to navigate to an outdated target position and fail the final subtask.}
	\label{ex-dynamic-quali} 
    \vspace{-0.4cm}
\end{figure*}

\textbf{Ablation study:}
We conduct ablation studies to evaluate the VLM backend, the exploration--exploitation transition, and the dual-source confirmation design. As shown in Table~\ref{tab:fine_ablation}, replacing GPT-4o with Llama3.2-Vision reduces SR to 25.6\%, indicating that stronger vision-language reasoning benefits fine-grained grounding, while the framework remains usable with an alternative VLM. Without the exploration--exploitation transition, SR decreases slightly
to 26.5\%, suggesting a modest benefit in final commitment robustness. The largest degradation occurs when egocentric VLM evidence is removed:
the system then relies only on memory-based cues, and SR drops to 23.1\%. 
For fine-grained visual and relational query matching, the egocentric VLM observation subgraph complements the accumulated memory subgraph.

\subsection{Dynamic Long-Horizon Object Navigation}
To evaluate dynamic scene update and long-horizon navigation efficiency, we construct dynamic object navigation tasks in the Habitat simulator using scenes from HM3D~\cite{ramakrishnan2021habitat} and Gibson~\cite{xia2018gibson}. Specifically, we selected four indoor scenes from each dataset. In each scene, the robot is required to complete a sequence of four object navigation subtasks, where the target objects are specified either by semantic-level descriptions, such as \textit{toilet}, or by instance-level descriptions, such as \textit{bed with a blue comforter}.

All four target objects are initially present at the beginning of each long-horizon task. To simulate scene dynamics, after the first subtask is completed, the target object of the fourth subtask is moved to a different location through simulator manipulation. This setting creates a stale-memory challenge: a robot that relies on the originally mapped object location may navigate to an outdated target position when the fourth subtask is executed. 

We compare OmniNav with UniGoal\cite{yin2025unigoal} and DualMap\cite{jiang2025dualmap}, whose original implementations are adapted to our long-horizon multi-object navigation protocol in
previously unknown environments. Under this setting, all methods must explore the environment online and sequentially navigate to multiple target
objects without access to a pre-built map. We additionally compare OmniNav with OmniNav*, a variant without dynamic memory updating. In the final subtask, OmniNav*
retrieves a candidate matching the target query from its previously constructed map, navigates to the retrieved candidate location, and stops upon arrival. Consequently, when the target has moved, OmniNav* may still navigate to its stale location, allowing us to isolate the effect of dynamic memory
  updating. All methods use identical observation modalities, per-subtask step
budgets, and scene-change schedules.
For each task, we average subtask SR and SPL over three trials with
different starting positions to evaluate the full sequence.

\textbf{Quantitative results:} Fig.~\ref{dynamic-ex} reports the subtask-level SR and SPL on dynamic long-horizon object navigation. UniGoal performs relatively well at the beginning of the sequence, but its average SR decreases markedly in later subtasks, while DualMap exhibits consistently low and fluctuating performance. 
In our map-from-scratch setting, DualMap exhibits lower
robustness under sequential target switches. This performance gap may partly reflect the difficulty of estimating anchors and anchor--object associations from partial observations, together with the sensitivity of maximum-similarity retrieval to semantically related distractors. These challenges become more pronounced when scene knowledge must be incrementally constructed and revised.

For OmniNav and OmniNav* alike, persistent memory enables observations from earlier subtasks to be reused for later targets, improving navigation efficiency over the sequence. The benefit of dynamic memory updating is most evident after the final target is relocated. OmniNav* retains the stale target location and therefore often terminates at an outdated candidate, leading to degraded SR and SPL in the fourth subtask. In contrast, OmniNav invalidates the stale target instance and resumes
retrieval or exploration, achieving higher
fourth-subtask SR and SPL.

\textbf{Qualitative results:}
Fig.~\ref{ex-dynamic-quali} visualizes two representative dynamic long-horizon object navigation episodes. In each episode, the robot sequentially visits multiple target objects, while the final target is relocated during execution. OmniNav and OmniNav* follow the same task
protocol and scene-change schedule, and exhibit similar trajectories before the final subtask. The difference emerges when the robot searches for the relocated final target. OmniNav identifies the stale instance and
updates the active memory before re-localizing the moved target, whereas OmniNav* continues to rely on outdated memory and fails on the final subtask. These examples qualitatively illustrate the benefit of online
scene memory revision under target relocation.

\begin{figure}[t!]\centering
	\includegraphics[width=7cm]{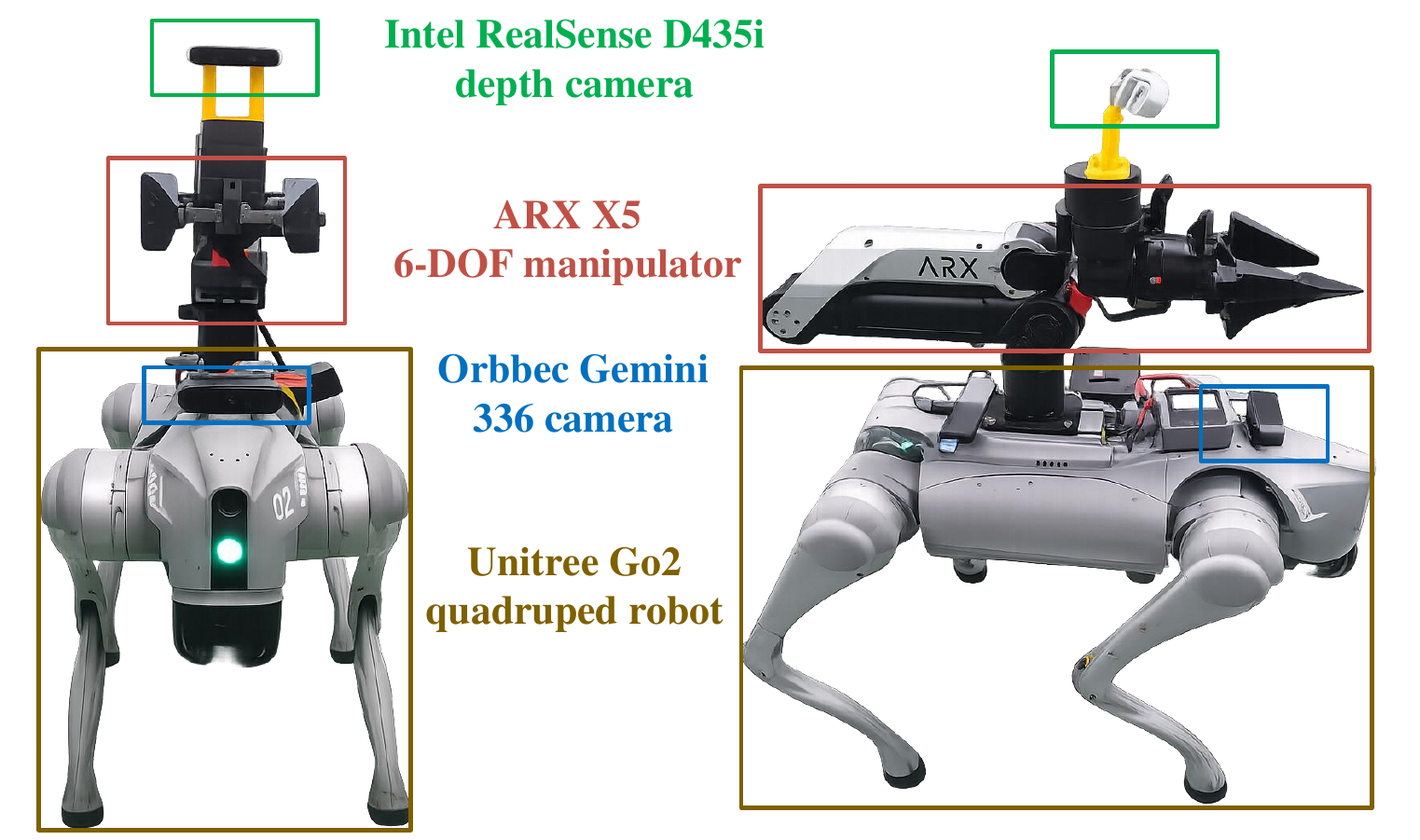}
	\caption{
Mobile manipulation platform used in the real-world experiments.
The platform consists of a Unitree Go2 quadruped robot equipped with an ARX X5 6-DOF manipulator, a Unitree 4D LiDAR L2 for robot pose estimation, an Orbbec Gemini 336 head-mounted RGB-D camera for perception and navigation, and an Intel RealSense D435i wrist-mounted depth camera for manipulation feedback.
}
	\label{ex-robot} 
    \vspace{-0.4cm}
\end{figure}

\section{Real-world Experiments}
\label{real-Exp}
In the real-world experiments, we evaluate OmniNav on two representative embodied tasks: long-horizon navigation in dynamic environments and long-horizon mobile manipulation with object pick-and-place operations.

\textbf{Robot hardware:} 
We evaluate OmniNav on a Unitree Go2 quadruped robot equipped with an ARX X5 6-DOF manipulator (Fig.~\ref{ex-robot}). The platform uses a Unitree 4D LiDAR L2 for pose estimation, an Orbbec Gemini 336 head-mounted RGB-D camera for perception and navigation, and an Intel RealSense D435i wrist camera for manipulation perception. An onboard NVIDIA GeForce RTX 3090 handles perception, mapping, navigation, and local inference, while GPT-4o-based LLM/VLM inference is accessed via API.

\begin{figure*}[t!]\centering
	\includegraphics[width=16.5cm]{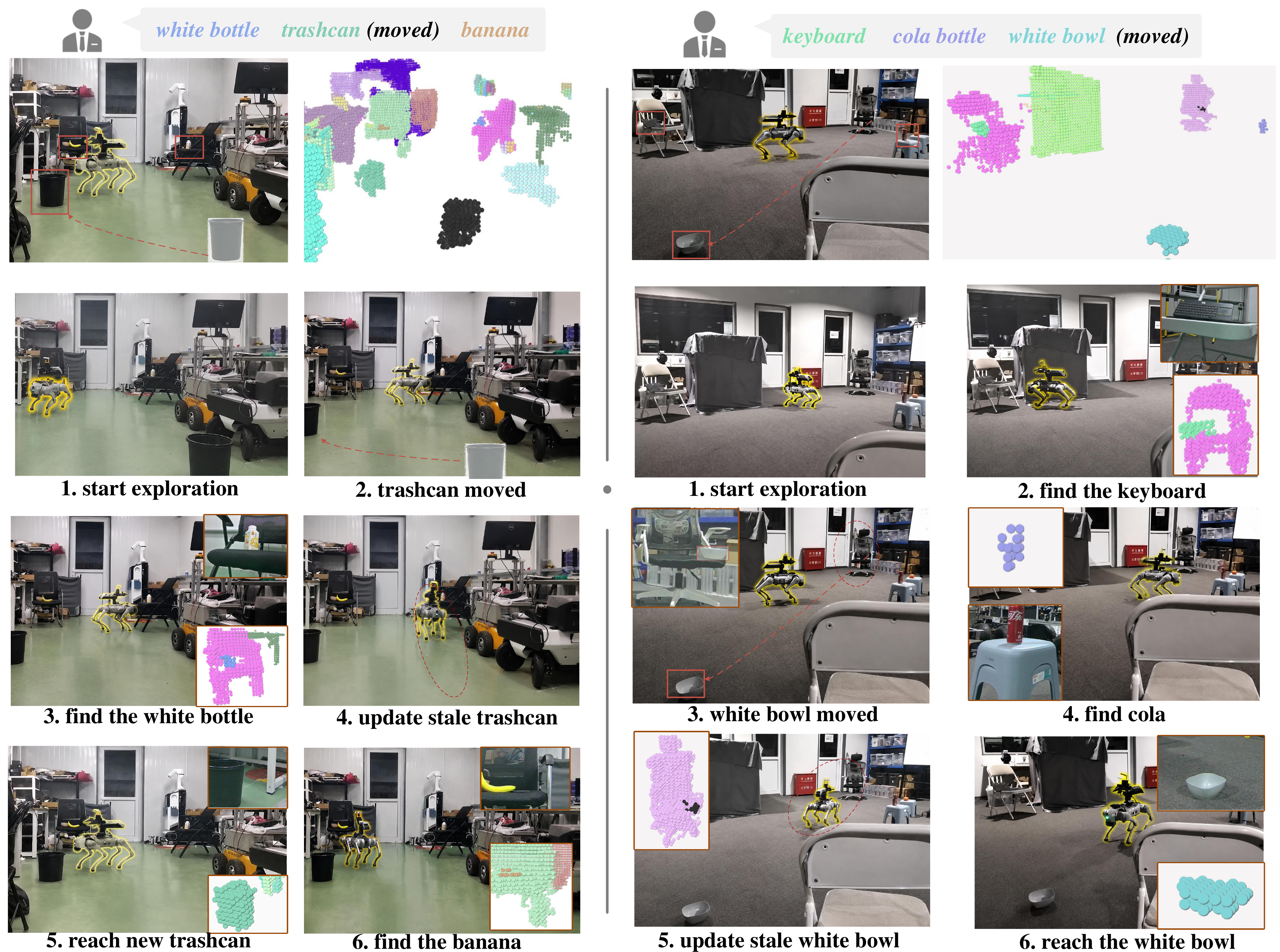}
	\caption{Qualitative results of dynamic long-horizon object navigation in real-world environments.
Each task requires the robot to sequentially navigate to multiple target objects, while one target is deliberately relocated during execution.
OmniNav removes stale object entries from its online object-centric
memory, updates the scene representation, and resumes target search
using the revised memory.}
	\label{real-dynamic} 
    \vspace{-0.3cm}
\end{figure*}

\subsection{Dynamic Long-Horizon Object Navigation}
\textbf{Environments and tasks:} We conduct real-world long-horizon object navigation experiments in three dynamic indoor scenarios: a laboratory, a home office, and a spacious corridor. In each scenario, the robot is required to sequentially navigate to three objects of different categories. During execution, one target is deliberately relocated, creating
a stale-memory challenge for the maintained object-centric scene memory. Each task is repeated three times with different initial robot positions. We report the number of successful trials for each sequential target in every scenario.

\begin{table}[t]
\centering
\caption{Real-world dynamic long-horizon object navigation results under target relocation.
Each subtask is evaluated over three trials with different initial robot positions.}
\label{tab:real_dynamic_nav}
\resizebox{0.8\linewidth}{!}{
\begin{tabular}{lccc}
\toprule
\textbf{Scene} & \textbf{Subtask 1} & \textbf{Subtask 2} & \textbf{Subtask 3} \\
\midrule
Corridor    & 3/3  & 3/3  & 2/3  \\
Laboratory  & 3/3  & 3/3  & 3/3  \\
Home office & 3/3  & 3/3  & 3/3  \\
\bottomrule
\end{tabular}
}
\end{table}

\textbf{Quantitative results:} Table~\ref{tab:real_dynamic_nav} reports the trial-level success results in three real-world
dynamic scenarios. OmniNav succeeds in all trials for the first two subtasks and in 8 of 9 trials for the relocated target. 
These results indicate the feasibility of online scene memory revision
in the three evaluated real-world scenarios under target relocation.

\textbf{Qualitative results:}
Fig.~\ref{real-dynamic} shows two representative real-world dynamic long-horizon navigation episodes. \mbox{OmniNav} maintains object-centric scene memory across sequential subtasks. When a previously observed target is relocated, OmniNav invalidates
and removes the instance associated with its former location, preventing
subsequent navigation from being guided by stale target information. The robot then re-enters target search using the updated memory and reaches the valid target. These examples highlight the importance of online scene memory maintenance for reliable long-horizon navigation in changing environments.

\subsection{Long-Horizon Mobile Manipulation Tasks}

\textbf{Environments and tasks:} We evaluate long-horizon mobile manipulation in two indoor scenes, a laboratory and a home office. Following the long-horizon OVMM protocol~\cite{yenamandra2023homerobot}, each trial requires the robot to locate and grasp a target object, navigate to a designated receptacle, and place the object.

To evaluate robustness under different execution conditions, we design three difficulty levels. \textbf{Level 1 (Direct)} places the target object within the robot's initial field of view, requiring little navigation before interaction. \textbf{Level 2 (Navigation)} places the target initially out of sight, requiring exploration, target navigation, and global planning. \textbf{Level 3 (Disturbed)} builds on Level 2 by introducing unannounced perturbations during execution, such as relocating the target when the robot approaches the original location of the target. We conduct 20 trials for each level, resulting in 60 trials in total.

\textbf{Baselines:}
Because few comparable systems provide reproducible implementations for our hardware setting, we adopt the open-source OK-Robot~\cite{liu2024okrobot} as an adapted baseline, denoted as OK-Robot*. OK-Robot* uses the same perception and navigation stack as OmniNav, but follows a decoupled, open-loop execution pipeline. After reaching the navigation endpoint, it performs a single-shot grasp attempt without
reachability-aware base-pose refinement or closed-loop recovery.

\textbf{Metrics:} We report three metrics: 
(1) Success Rate (SR), the percentage of trials in which the robot completes the full pick-and-place sequence; 
(2) Step-wise Success Rate (SSR), the
conditional success of each execution stage among trials reaching that stage; and
(3) \mbox{Recovery Rate (RR)}, the proportion of detected execution anomalies successfully resolved through hierarchical recovery without task termination.

\begin{table}[t]
\centering
\caption{Success rates (SR) across different mobile manipulation difficulty levels (20 trials
each).}
\label{MM-level_success}
\resizebox{0.8\linewidth}{!}{
\begin{tabular}{lcccc}
\toprule
\textbf{Method} & \textbf{Level 1} & \textbf{Level 2} & \textbf{Level 3} & \textbf{Overall} \\
\midrule
OK-Robot$^\star$\cite{liu2024okrobot} & \cellcolor[HTML]{FFD8B2}80.0\% & \cellcolor[HTML]{FFD8B2}55.0\% & \cellcolor[HTML]{FFD8B2}25.0\% & \cellcolor[HTML]{FFD8B2}53.3\% \\
\textbf{OmniNav} & \cellcolor[HTML]{FFB2B2}85.0\% & \cellcolor[HTML]{FFB2B2}75.0\% & \cellcolor[HTML]{FFB2B2}55.0\% & \cellcolor[HTML]{FFB2B2}71.7\% \\
\bottomrule
\end{tabular}
}
\end{table}


\begin{table}[t]
\centering
\caption{Overall performance comparison and ablation study for real-world long-horizon mobile manipulation.}
\label{MM-overall_results}
\resizebox{0.6\linewidth}{!}{
\begin{tabular}{lccc}
\toprule
\textbf{Method} & \textbf{SR$\uparrow$} & \textbf{RR$\uparrow$}   \\
\midrule
OK-Robot$^\star$\cite{liu2024okrobot}  & 53.3\% & --    \\
\noalign{\vskip 2pt}
\hdashline[5pt/1pt]
\noalign{\vskip 3pt}
\textbf{OmniNav} & \cellcolor[HTML]{FFB2B2}71.7\% & \cellcolor[HTML]{FFB2B2}71.4\%   \\
w/o Base Alignment & \cellcolor[HTML]{FFD8B2}60.0\% & \cellcolor[HTML]{FFD8B2}60.0\%  \\
w/o Hier. Recovery & \cellcolor[HTML]{FFFFB2}55.0\% & --  \\
\bottomrule
\end{tabular}
}
\end{table}

\begin{table}[t]
\centering
\caption{Step-wise Success Rate (SSR) breakdown for real-world mobile manipulation.}
\label{MM-ssr_breakdown}
\resizebox{0.75\linewidth}{!}{
\begin{tabular}{lccc}
\toprule
\textbf{Method} & \textbf{Alignment}  & \textbf{Grasp} & \textbf{Place} \\
\midrule
OK-Robot$^\star$\cite{liu2024okrobot}  & -- & \cellcolor[HTML]{FFD8B2}69.2\% & \cellcolor[HTML]{FFD8B2}88.9\% \\
\textbf{OmniNav}  & 92.8\% & \cellcolor[HTML]{FFB2B2}80.0\% & \cellcolor[HTML]{FFB2B2}100.0\% \\
\bottomrule
\end{tabular}
}
\end{table}

\begin{figure*}[t!]\centering
	\includegraphics[width=17cm]{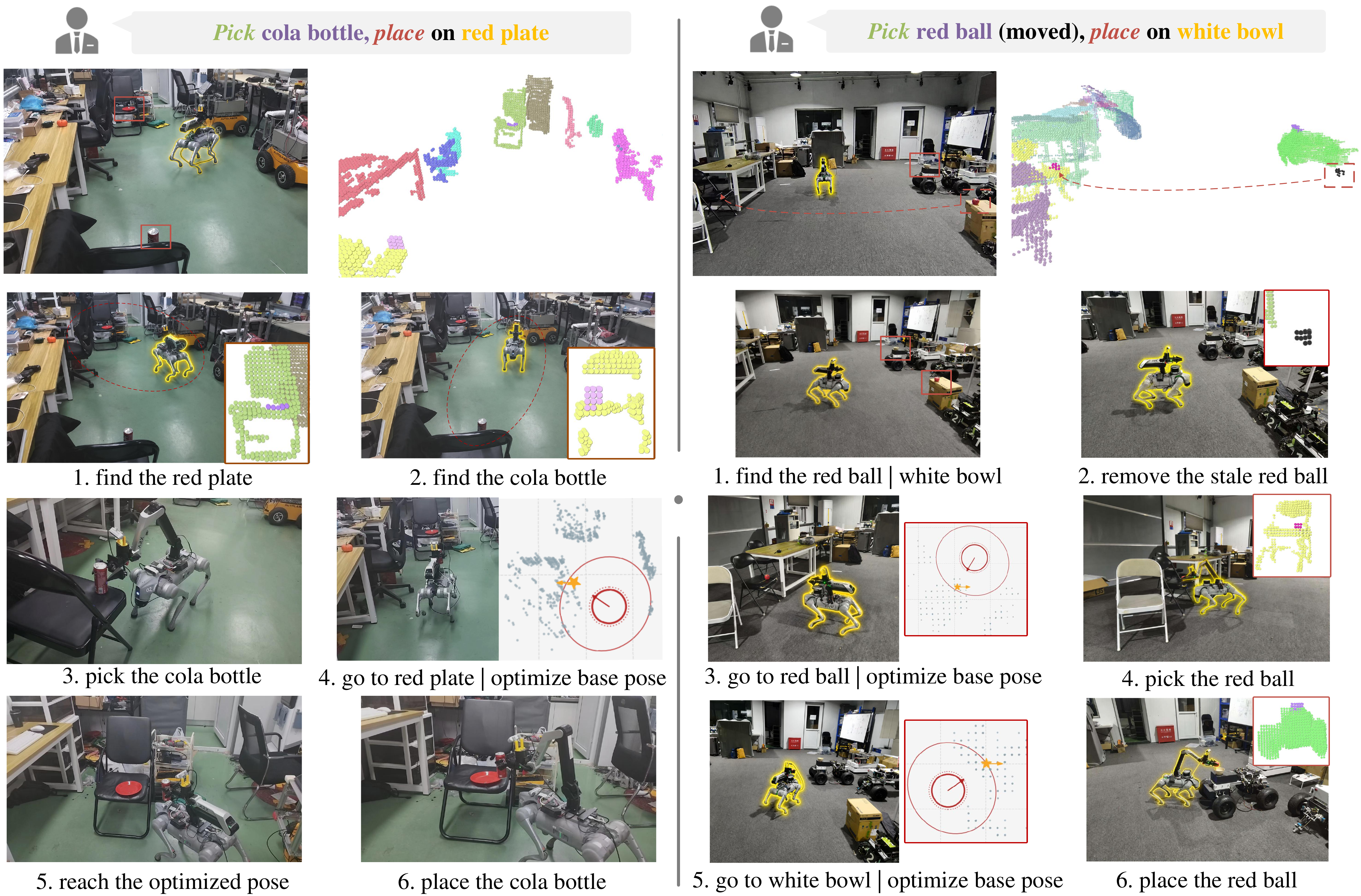}
	\caption{
Qualitative real-world long-horizon mobile manipulation results. Left:
OmniNav completes a pick-and-place task by searching for the target,
grasping it, navigating to the receptacle, selecting a reachability-aware
base pose, and placing it. Right: In a more challenging scenario, OmniNav handles target relocation by updating the scene memory,
re-finding the moved target, and continuing the task.
}
	\label{real-dynamic-MM} 
    \vspace{-0.4cm}
\end{figure*}

\textbf{Quantitative results:} Table~\ref{MM-level_success} reports the success rates across different difficulty levels. At Level 1, where the target is within the robot's initial field of view and no long-range navigation is required, both methods achieve comparable performance. This is expected because the short interaction distance leaves limited room for reachability-aware alignment or recovery mechanisms to provide additional benefit. At Level 2, OmniNav improves the success rate by 20 percentage points over OK-Robot*, suggesting that navigation--manipulation inconsistency becomes an important bottleneck once target search and long-range navigation are required. At Level 3, where unannounced perturbations are introduced during execution, the performance gap further increases to 30 percentage points. Since OK-Robot* follows an open-loop pipeline, mid-task perturbations often lead to immediate failure. In contrast, OmniNav handles many of these anomalies through hierarchical recovery, substantially improving the robustness of the task.

Table~\ref{MM-overall_results} further summarizes the overall SR and RR. Removing either reachability-aware base alignment or hierarchical recovery leads to a clear decline in overall success rate, highlighting the importance of both components.

We also report the Step-wise Success Rate (SSR) in Table~\ref{MM-ssr_breakdown}. Since OK-Robot* and OmniNav share the same perception and navigation stack, the breakdown focuses on the execution stages after target localization. The stage-wise breakdown shows 92.8\% alignment success for OmniNav,
with higher grasp (80.0\% vs. 69.2\%) and place
(100.0\% vs. 88.9\%) success than OK-Robot*. These results suggest that improved navigation-to-manipulation
coordination is an important contributor to the overall performance gain, with reachability-aware base alignment and closed-loop recovery helping convert target localization into successful physical interaction.

\textbf{Qualitative results:} Fig.~\ref{real-dynamic-MM} shows real-world long-horizon mobile manipulation tasks.
OmniNav searches for the target object, maintains an object-centric
scene memory, and refines the navigation endpoint to an
interaction-feasible base pose. This refinement
helps translate target localization into physical interaction. In the
dynamic case, when the target is moved during execution, OmniNav
invalidates the stale target instance, updates the scene memory, and
re-initiates target search and confirmation before resuming
manipulation. These examples illustrate how persistent scene memory,
reachability-aware base-pose refinement, and closed-loop recovery
jointly support robust long-horizon pick-and-place execution.

\begin{table}[t]
    \centering
    \caption{Runtime of the main online and offline modules.}
    \label{tab:runtime}
    \small
    \setlength{\tabcolsep}{3.5pt}
    \renewcommand{\arraystretch}{1.12}

    \resizebox{0.8\linewidth}{!}{%
    \begin{tabular}{
        @{}
        l
        S[table-format=4.2]
        l
        @{}
    }
        \toprule
        \textbf{Module}
        & {\textbf{Runtime (s)}}
        & \textbf{Invocation} \\
        \midrule

        \multicolumn{3}{@{}l}{\textit{Online modules}} \\
        \addlinespace[1pt]

        Scene memory update
        & 0.30
        & per frame \\

        Frontier selection
        & 0.52
        & per decision \\

        Target confirmation
        & 4.58
        & per candidate \\

        Base-pose optimization
        & 2.85
        & per call \\
        \addlinespace[1pt]
        \hdashline[5pt/1pt]
        
        \addlinespace[3pt]
        \multicolumn{3}{@{}l}{\textit{Offline preprocessing}} \\
        \addlinespace[1pt]

        Co-occurrence prior construction
        & 15.71
        & one-time \\

        Reachability-shell construction
        & 1125.40
        & one-time \\

        \bottomrule
    \end{tabular}%
    }
\end{table}

\subsection{Runtime Analysis}
Table~\ref{tab:runtime} summarizes the computational cost of
OmniNav. Among the online modules, scene memory update, \mbox{including}
object recognition, instance segmentation, and 3D map fusion, takes
0.30\,s per frame, while frontier selection requires 0.52\,s per
decision. Target confirmation takes 4.58\,s per candidate, and
base-pose optimization takes 2.85\,s per call. Frontier selection is performed intermittently, while target confirmation and base-pose optimization are invoked only upon candidate observation and physical interaction, respectively. 
Thus, only the scene-memory update runs per frame, whereas the more expensive decision-level modules incur intermittent or event-triggered costs, supporting online long-horizon execution.

The offline co-occurrence prior construction and reachability-shell
construction require 15.71\,s and 1125.40\,s, respectively. The latter
is dominated by inverse-kinematics evaluation over the sampled
workspace points, and its runtime therefore scales approximately with
the number of sampled points. Although this preprocessing is
comparatively expensive, it is performed only once for a given manipulator model, and the fitted shell can be reused across subsequent tasks without additional online overhead.

\section{Conclusion} \label{C}

In this work, we present OmniNav, a training-free framework for robust long-horizon target navigation and interaction in dynamic indoor environments. OmniNav formulates long-horizon execution as continual inference over a factorized posterior coupling scene validity, target belief, and interaction feasibility.
It maintains an updatable object-centric scene memory with Bayesian persistence inference, revises region-level target beliefs using semantic context and unsuccessful exploration, and verifies fine-grained candidates by combining accumulated scene context with egocentric VLM observations. For physical interaction, OmniNav further refines navigation endpoints according to manipulation reachability and collision safety, while hierarchical recovery propagates execution feedback to subsequent decisions.
Experiments across navigation benchmarks, dynamic long-horizon tasks, and real-world mobile manipulation demonstrate efficient target search, instance-level identification, adaptation to scene changes, and successful physical interaction.

OmniNav nevertheless has several limitations. Its scene memory quality and fine-grained confirmation depend on the accuracy of visual-language perception, depth sensing, and camera-pose estimation. Scene changes can only be detected after the affected region becomes observable again, which may delay memory revision in large or sparsely revisited environments. In addition, the recovery mechanism covers only a limited set of manipulation failures and target-relocation events.
Future work will investigate active change detection and broader
recovery strategies across embodied tasks.

\bibliographystyle{Bibliography/IEEEtran}
\bibliography{Bibliography/mycite}

\end{document}